\documentclass[10pt,journal,compsoc]{IEEEtran}

\usepackage{tikz}
\usepackage{pgfplots}
\usepackage{xcolor}
\usepackage{filecontents}
\usepackage{tablefootnote}
\usepackage{amsmath}
\usepackage{graphicx}
\usepackage{url}
\usepackage{hyperref}
\usepackage{subfig}
\usepackage{amsfonts}
\usepackage{color, soul}
\usepackage{mathrsfs}
\usepackage{cleveref}
\usepackage{multirow}
\usepackage{float}
\usepackage{wrapfig}
\usepackage{amsthm}
\usepackage{bm}
\usepackage{bbm}
\usepackage{algorithm}
\usepackage{algorithmic}
\usepackage{colortbl}
\usepackage{enumitem}
\usepackage{color}
\usepackage{balance}
\usepackage{stfloats}
\usepackage{makecell}

\ifCLASSOPTIONcompsoc
  \usepackage[nocompress]{cite}
\else
  \usepackage{cite}
\fi
\ifCLASSINFOpdf
\else
\fi

\usepackage{multirow}
\usepackage{amsmath}
 \usepackage{amsfonts,amssymb}
\usepackage{bbm}
\begin{document}

%
\title{Fairness in Graph Mining: A Survey}

\author{Yushun Dong, Jing Ma, Song Wang, Chen Chen, and Jundong Li
\IEEEcompsocitemizethanks{
\IEEEcompsocthanksitem Y. Dong is with Department of Electrical and Computer Engineering, University of Virginia, Charlottesville, Virginia, US.\protect\\
E-mail: yd6eb@virginia.edu
\IEEEcompsocthanksitem J. Ma is with Department of Computer Science, University of Virginia, Charlottesville, Virginia, US.\protect\\
E-mail: jm3mr@virginia.edu
\IEEEcompsocthanksitem S. Wang is with Department of Electrical and Computer Engineering, University of Virginia, Charlottesville, Virginia, US.\protect\\
E-mail: sw3wv@virginia.edu
\IEEEcompsocthanksitem C. Chen is with Biocomplexity Institute, University of Virginia, Charlottesville, Virginia, US.\protect\\
 E-mail: zrh6du@virginia.edu
\IEEEcompsocthanksitem J. Li is with Department of Electrical and Computer Engineering, Department of Computer Science, and School of Data Science, University of Virginia, Charlottesville, Virginia, US.\protect\\
E-mail: jundong@virginia.edu
}
}

%
%

\markboth{IEEE TRANSACTIONS ON KNOWLEDGE AND DATA ENGINEERING}%
{Shell \MakeLowercase{\textit{et al.}}: Bare Advanced Demo of IEEEtran.cls for IEEE Computer Society Journals}
%



\IEEEtitleabstractindextext{
\begin{abstract}

Graph mining algorithms have been playing a significant role in myriad fields over the years. However, despite their promising performance on various graph analytical tasks, most of these algorithms lack fairness considerations.
As a consequence, they could lead to discrimination towards certain populations when exploited in human-centered applications.
Recently, algorithmic fairness has been extensively studied in graph-based applications.
In contrast to algorithmic fairness on independent and identically distributed (i.i.d.) data, fairness in graph mining has exclusive backgrounds, taxonomies, and fulfilling techniques.
In this survey, we provide a comprehensive and up-to-date introduction of existing literature under the context of fair graph mining.
Specifically, we propose a novel taxonomy of fairness notions on graphs, which sheds light on their connections and differences.
We further present an organized summary of existing techniques that promote fairness in graph mining. Finally, we discuss current research challenges and open questions, aiming at encouraging cross-breeding ideas and further advances. 
\end{abstract}

\begin{IEEEkeywords}
Algorithmic Fairness, Graph Mining, Debiasing
\end{IEEEkeywords}}

\maketitle

\IEEEdisplaynontitleabstractindextext

%
\IEEEpeerreviewmaketitle

\ifCLASSOPTIONcompsoc
\IEEEraisesectionheading{\section{Introduction}\label{sec:introduction}}
\else
\section{Introduction}
\label{sec:introduction}
\fi
\label{intro}

Graph-structured data is pervasive in diverse real-world applications, e.g., E-commerce~\cite{niu2020dual,li2020hierarchical}, health care~\cite{cui2020deterrent,fritz2021combining}, traffic forecasting~\cite{jiang2021graph,li2017diffusion}, and drug discovery~\cite{bongini2021molecular,xiong2021graph}.
In recent years, a number of graph mining algorithms have been proposed to gain a deeper understanding of such data.
These algorithms have shown promising performance on graph analytical tasks such as node classification~\cite{DBLP:conf/iclr/KipfW17,grover2016node2vec,velivckovic2017graph} and link prediction~\cite{lu2011link,liben2007link,al2006link}, contributing to great advances in many graph-based applications.

Despite the success of these graph mining algorithms, most of them lack fairness considerations.
Consequently, they could yield discriminatory results towards certain populations when such algorithms are exploited in human-centered applications~\cite{kang2021fair}.
%
For example, a social network-based job recommender system may unfavorably recommend fewer job opportunities to individuals of a certain gender~\cite{lambrecht2019algorithmic} or individuals in an underrepresented ethnic group~\cite{sweeney2013discrimination}. 
With the widespread usage of graph mining algorithms, such potential discrimination could also exist in other high-stake applications such as disaster response~\cite{van2020automated}, criminal justice~\cite{agarwal2021towards}, and loan approval~\cite{sarkar2020mitigating}.
%
%
In these applications, critical and life-changing decisions are often made for the individuals involved.
Therefore, how to tackle unfairness issues in graph mining algorithms naturally becomes a crucial problem.

\begin{figure}[!t]
    \vspace{-3.5mm}
    \centering
    \subfloat[Unbiased graph topology]{
        \includegraphics[width=0.225\textwidth]{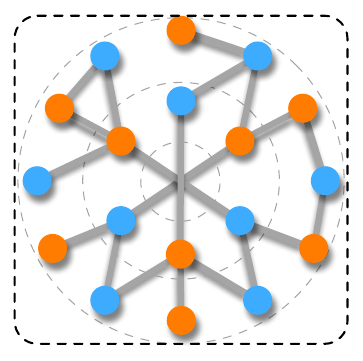}
        \label{dis_1_1}
    }
    \subfloat[Biased graph topology]{
        \includegraphics[width=0.219\textwidth]{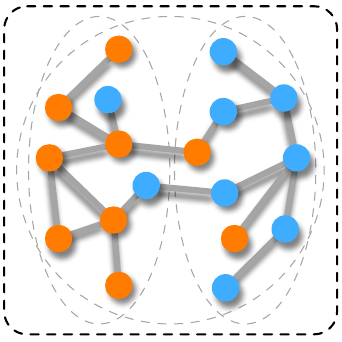}
        \label{st_1}
    }  \\
    \vspace{-2mm}
        \subfloat[An example of biased node embeddings (learned via information propagation mechanism of GNNs) induced by biased input graph.]{
        \includegraphics[width=0.452\textwidth]{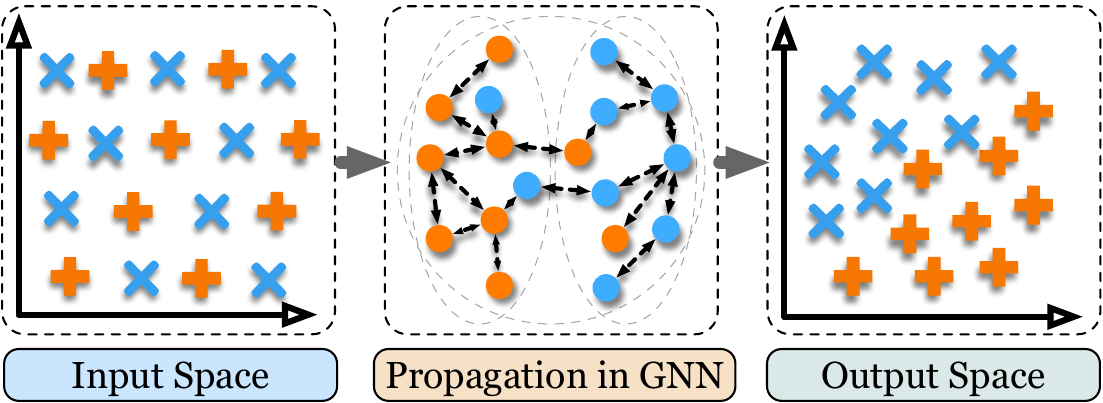}
        \label{st_2}
    }
    \vspace{-2mm}
    \caption{Examples of (a) unbiased graph topology, (b) biased graph topology, and (c) how information propagation mechanism induces bias in GNNs. Nodes in two different demographic subgroups are in orange and blue.}
        \vspace{-7mm}
    \label{bias_incorporating}
\end{figure}

Compared with achieving fairness in the context of independent and identically distributed (i.i.d.) data, fulfilling fairness in graph mining can be non-trivial due to two main challenges.
The first challenge is to formulate proper fairness notions as the criteria to determine the existence of unfairness (i.e., bias).
Although a vast amount of traditional algorithmic fairness notions have been proposed centered on i.i.d. data~\cite{mehrabi2021survey,du2020fairness}, they are unable to reflect the bias exhibited by the relational information (i.e., the topology) in graph data.
%
%
For example, the same population can be connected with different topologies as in Fig.~\ref{dis_1_1} and~\ref{st_1}, where each node represents an individual, and the color of nodes denotes their demographic subgroup membership, such as different genders. 
Compared with the graph topology in Fig.~\ref{dis_1_1}, the topology in Fig.~\ref{st_1} has more intra-group edges than inter-group edges.
The dominance of intra-group edges in the graph topology is a common type of bias existing in real-world graphs~\cite{dai2021say,dong2021edits,jalali2020information}, which cannot be captured by traditional algorithmic fairness notions.
The second challenge is to prevent the graph mining algorithms from inheriting the bias exhibited in the input relational information~\cite{varshney2019pretrained,stoica2018algorithmic,ma2021subgroup,dong2021edits}.
%
%
%
%
We present a toy example to demonstrate how the information propagation mechanism in Graph Neural Networks (GNNs)~\cite{DBLP:conf/iclr/KipfW17,hamilton2017inductive,velivckovic2017graph} induces bias to the output node embeddings from a biased graph topology in Fig.~\ref{st_2}.
In the input space, the node features are uniformly distributed. However, when the information propagation is performed on a biased topology as in Fig.~\ref{st_1}, the information received by nodes in different subgroups could be biased~\cite{dong2021edits}, leading to a biased embedding distribution in the output space.

There has been emerging research interest in fulfilling algorithmic fairness in graph mining. Nevertheless, the studied fairness notions vary across different works, which can be confusing and impede further progress.
Meanwhile, different techniques are developed in achieving various fairness notions. Without a clear understanding of the corresponding mappings, future fair graph mining algorithm design can be difficult.
%
Therefore, a systematic survey of recent advances is needed to shed light on future research.
In this survey, we present a comprehensive and up-to-date review of existing works in fair graph mining.
%
%
The main contributions of this survey paper are summarized as:

\begin{itemize}[topsep=0pt]
    \item \textbf{\emph{Novel Fairness Taxonomy.}} We propose a novel taxonomy of fairness notions in graph mining. Such a taxonomy includes five groups of fairness notions: group fairness, individual fairness, counterfactual fairness, degree-related fairness, and application-specific fairness. For each group of fairness notions, we present their definitions and common quantitative metrics.
    \item \textbf{\emph{Comprehensive Technique Review.}} We provide a comprehensive and organized review of six groups of techniques that are commonly utilized to promote fairness in graph mining algorithms. For each group of techniques, we summarize representative formulations under different fairness notions.
    \item \textbf{\emph{Rich Public-Available Resources.}} We collect rich resources of algorithms and benchmark datasets that can be employed for fair graph mining research. Therefore, this survey can facilitate the development of new graph mining approaches to promote fairness.
    \item \textbf{\emph{Challenges and Future Directions.}} We present the limitations of current research and point out pressing challenges. Open research questions are also discussed for further advances.
\end{itemize}

\noindent \textbf{Difference from Existing Surveys.}
Despite the urgent need for a systematic overview of algorithmic fairness in graph mining, most existing related survey papers are under the context of i.i.d. data~\cite{pessach2020algorithmic,mehrabi2021survey,du2020fairness,caton2020fairness,corbett2018measure,mitchell2021algorithmic}.
A few other surveys pay attention to the algorithmic fairness in relational data~\cite{pitoura2021fairness,zhang2022fairness}. Nevertheless, they are limited to either a certain application scenario (e.g., recommender systems) or a certain type of graph mining algorithm (e.g., machine learning-based algorithms).
As a consequence, there still lacks an inclusive overview of the fairness notions and fulfilling techniques for different graph mining algorithms.
This serves as the primary motivation for this survey.
Different from the survey papers above, this survey includes: (1) a systematic review of existing fairness notions in the realm of graph mining; and (2) a well-organized introduction to commonly used fairness-fulfilling techniques for various graph mining algorithms.



\noindent \textbf{Intended Audiences.}
The intended audiences of this survey are (1) researchers who would like to understand how fairness is defined and fulfilled in graph mining; and (2) practitioners who plan to generalize fair graph mining approaches to different applications.

\noindent \textbf{Survey Structure.}
The remainder of this survey paper is organized as follows.
Section~\ref{labels_section} introduces the notations and preliminaries.
In Section~\ref{notion_section}, different fairness notions and corresponding metrics are systematically reviewed. 
Based on these fairness notions and metrics, Section~\ref{technique_section} introduces six groups of techniques to fulfill fairness.
%
%
Section~\ref{challenge_and_lib} discusses existing research challenges and open questions.
%
Finally, Section~\ref{conclusion_section} presents the conclusion of this survey.


\begin{figure*}[!t]
    \centering
    \includegraphics[width=0.995\textwidth]{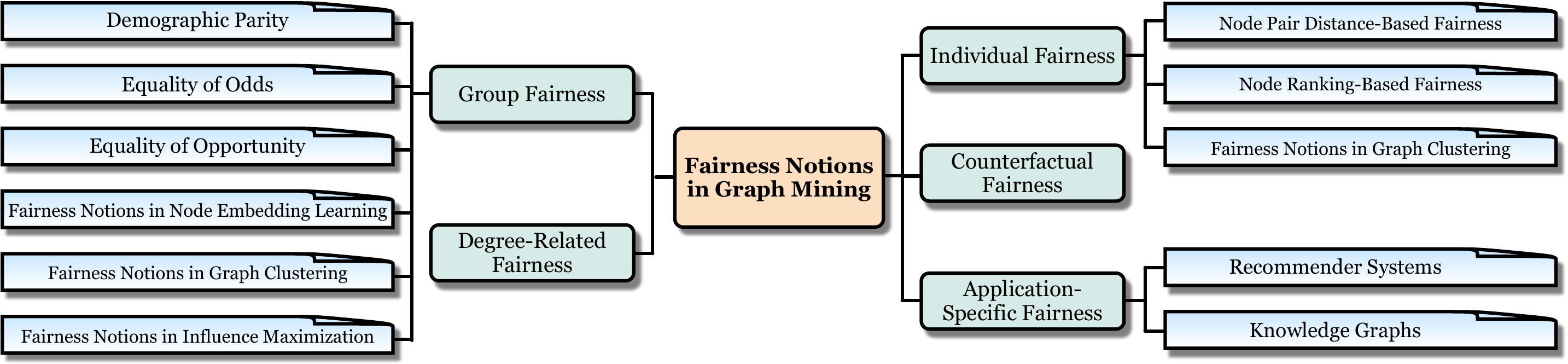}
    \vspace{-2mm}
    \caption{Taxonomy of algorithmic fairness notions in graph mining algorithms.}
    \vspace{-6mm}
    \label{taxonomy}
\end{figure*}

\section{Notations and Preliminaries}
\label{labels_section}

\begin{table}[t!]
\caption{Notations and the corresponding descriptions.} 
\vspace{-0.2cm}
\label{tb:symbols}
\small
\begin{tabular}{c|l}
\hline
\textbf{Notations}       & \textbf{Definitions or Descriptions} \\
\hline
$| \cdot |$  & Cardinality operator for any set. \\
$E[ \cdot ]$ & Expectation operator.\\
$<\cdot, \cdot>$ & Inner product operator. \\
\hline
$\mathcal{G}$   &  The graph data. \\
$\mathcal{V}$   &  The set of nodes. \\
$\mathcal{E}$   &  The set of edges. \\
$\mathcal{X}$   &  The set of node features. \\
$\mathcal{A}$  & The seed set in influence maximization. \\
$\mathcal{N}_{v_i}$ & The one-hop neighboring node set of $v_i$.\\
$\mathcal{V}_i$ & The node set of the $i$-th sensitive subgroup.\\
\hline
$\mathbf{A}$            &    The adjacency matrix of graph $\mathcal{G}$.      \\
$\mathbf{A}^{\top}$            &    The transpose of adjacency matrix.       \\
$\mathbf{X}$            &    The node feature matrix of graph $\mathcal{G}$.      \\
$\mathbf{z}_i$ & The embedding of node $v_i$.\\
\hline
$v_i$ & The $i$-th node.\\
$n$    & The size of the node set $\mathcal{V}$.  \\
$d$    & The number of node features.  \\
$c$ & The class number for node classification.\\
\hline
\end{tabular}
\vspace{-1.0em}
\end{table}

In this section, we present important notations used throughout this paper and preliminaries. The notations and their definitions (or descriptions) are in Table~\ref{tb:symbols}.

\noindent \textbf{Notations.} We use bold uppercase letters (e.g., $\mathbf{A}$) and bold lowercase letters (e.g., $\mathbf{z}$) to denote matrices and vectors, respectively. For any matrix, e.g., $\mathbf{A}$, we refer to its ($i, j$)-th entry as $\mathbf{A}_{i,j}$, and the  transpose of $\mathbf{A}$ as $\mathbf{A}^\top$. We use letters in calligraphy font (e.g., $\mathcal{V}$) to denote sets. 

\noindent \textbf{Preliminaries.} 
In this paper, we use the terminology of ``graph" and ``network" interchangeably.
We denote a plain graph as $\mathcal{G}$ = ($\mathcal{V}$, $\mathcal{E}$), where $\mathcal{V}$ and $\mathcal{E}$ represent the set of nodes and edges, respectively. We use $\mathbf{A} \in \{0,1\}^{n \times n}$ to represent the graph adjacency matrix, where $n$ is the total number of nodes, and $\mathbf{A}_{i,j} = 1$ implies that there exists an edge between node $v_i$ and node $v_j$.
For graph data with node features (i.e., attributed networks), we use a matrix $\mathbf{X} \in \mathbb{R}^{n \times d}$ to denote the node feature matrix, where $d$ is the number of node features.
%
Unless otherwise specified, for the convenience of discussion on human-centered fairness, we consider that each individual is represented as a node in a graph by default. 
However, the literature in this survey and our discussion are not limited in this case. Additional clarifications will be presented in other scenarios.
%

%

\section{Fairness Notions in Graph Mining}
\label{notion_section}
In this section, we propose a novel taxonomy that categorizes existing fairness notions in graph mining into five groups, as presented in Fig.~\ref{taxonomy}. Following our proposed taxonomy, we hereby organize and introduce these fairness notions and their corresponding metrics.

\subsection{Group Fairness}

In many high-stake applications (e.g., loan approval systems~\cite{wang2019semi,xu2021towards}), certain features (e.g., race and gender) are protected by law to avoid being abused~\cite{mehrabi2021survey,du2020fairness,kamiran2012data,caton2020fairness}. Additionally, in applications such as online social networking, there are a number of features that users are usually unwilling to share (e.g., occupation and age)~\cite{dai2021say}.
These features are considered as protected or sensitive features~\cite{mehrabi2021survey}. Based on these features, the population can be divided into different demographic subgroups. Here we refer to these subgroups as sensitive subgroups.  
\textit{Group Fairness} is then defined upon such sensitive subgroups. Generally speaking, group fairness requires that the algorithm should not yield discriminatory predictions or decisions against individuals from any specific sensitive subgroup~\cite{DBLP:conf/innovations/DworkHPRZ12}. In this section, we introduce popular fairness notions under group fairness.

\subsubsection{Demographic Parity} 
\label{DP}

\textit{Demographic Parity} (a.k.a. \textit{Statistical Parity} and \textit{Independence}) is first introduced as a notion of group fairness based on binary sensitive feature(s) in binary classification tasks~\cite{DBLP:conf/innovations/DworkHPRZ12}. The binary sensitive feature divides the population into two sensitive subgroups (e.g., male/female). In binary classification tasks such as deciding whether a student should be admitted into a university or not, demographic parity is considered as achieved if the model yields the same acceptance rate for individuals in both sensitive subgroups. In graph mining, we first introduce demographic parity in node classification, followed by several extensions.

\noindent \textbf{Demographic Parity in Node Classification.} 
In node classification, we assume $\hat{Y}, S \in \{0, 1\}$ are the random variables representing the predicted class label and sensitive feature of a random node in the input graph, respectively.
The criterion of demographic parity is then formulated as
\begin{align}
\label{sp}
P(\hat{Y}=1 | S=0) = P(\hat{Y}=1 | S=1).
\end{align}
%
To quantify to what extent the demographic parity is satisfied, $\Delta_{DP}$ is defined when both the predicted labels and sensitive feature(s) are  binary~\cite{DBLP:journals/pvldb/LahotiGW19,dai2021say}. The formulation of $\Delta_{DP}$ is given as
\begin{align}
\Delta_{D P}=|P(\hat{Y}=1 \mid S=0)-P(\hat{Y}=1 \mid S=1)|.
\end{align}
The intuition here is to measure the acceptance rate difference between the two sensitive subgroups. 
However, the applicable scenarios can be limited if only binary sensitive attributes are considered.
In this regard, several following works extended demographic parity to multi-class sensitive feature scenarios~\cite{DBLP:conf/ijcai/RahmanS0019,spinelli2021biased}. The rationale is that the acceptance rates given by the algorithm should be the same across all sensitive subgroups.
To quantify demographic parity for multiple sensitive subgroups, Rahman et al.~\cite{DBLP:conf/ijcai/RahmanS0019} leveraged the variance of acceptance rates across all sensitive subgroups, while Spinelli et al.~\cite{spinelli2021biased} employed the largest acceptance rate difference among all subgroup pairs. 


\noindent \textbf{Extension to Link Prediction.}
In addition to the node classification task, demographic parity is also extended to the link prediction problem~\cite{laclau2021all}.
%
Specifically, we can obtain the average linking probability of node pairs spanning across different sensitive subgroups.
Then demographic parity is achieved when such probability is the same for any two pairs of sensitive subgroups~\cite{DBLP:conf/icml/BuylB20,woodworth2017learning}. 
Formally, assume ($i$, $j$) and ($k$, $l$) are the indices of two sensitive subgroup pairs (indices values can be the same within each tuple). The criterion of demographic parity is given as $ \delta_{i,j} = \delta_{k,l}, \,\,\forall i,j,k,l$,
%
where $\delta_{i,j}$ is the average linking probability of node pairs spanning across the $i$-th and the $j$-th sensitive subgroup, and it is formally defined as
\begin{align}
  \delta_{i,j} = \frac{1}{N_{i,j}}\sum_{v\in \mathcal{V}_i}\sum_{v'\in \mathcal{V}_j} P(f_{\text{link}} (v,v')=1).
  \label{eq:link_dp}
\end{align}
Here $P(f_{\text{link}} (v,v')=1)$ is the probability that the edge $(v,v')$ exists according to the link prediction model $f_{\text{link}}$; $\mathcal{V}_{i}$ and $\mathcal{V}_{j}$ represent the $i$-th and the $j$-th sensitive subgroup, respectively; $N_{i, j}$ is the number of node pairs spanning across $\mathcal{V}_{i}$ and $\mathcal{V}_{j}$, which is formally given as $N_{i, j}=\left|\left\{(v, v') \mid v \in \mathcal{V}_{i}, v' \in \mathcal{V}_{j}\right\}\right|$.
%
We then introduce $\Delta_{DP}^{\text{link}}$ that quantifies demographic parity in link prediction~\cite{DBLP:conf/icml/BuylB20,woodworth2017learning,saxena2021hm}. $\Delta_{DP}^{\text{link}}$ is defined as the largest absolute difference \cite{woodworth2017learning} between all pairs of $\delta_{i,j}$ and $\delta_{k,l}$, which is given as
\begin{equation}
    \Delta_{DP}^{\text{link}}=\max_{\forall i,j,k,l}|\delta_{i,j}-\delta_{k,l}|.
    \label{eq: delta_dp_link}
\end{equation}
Besides, a relaxed criterion of demographic parity has been defined by only focusing on intra- and inter-subgroup links~\cite{laclau2021all}.
%
%
Specifically, denote two random nodes as $v$ and $v'$, where $(v, v') \in \mathcal{V} \times \mathcal{V}$. Assume $s$ and $s'$ are the sensitive feature values of node $v$ and $v'$, respectively.
A relaxed criterion for demographic parity is defined as
\begin{align}
\label{link_pred_cri_2}
P(f_{\text{link}} (v,v')=1|s = s') = P(f_{\text{link}} (v,v')=1|s \neq s').
\end{align}
%
Generally, such a criterion requires that for a random pair of nodes, the probability that they are connected should be the same, regardless of whether their sensitive feature values are the same or not.
%
%
To quantify how well the criterion in Eq.~(\ref{link_pred_cri_2}) is satisfied, \textit{Disparate Impact} (DI) and \textit{Balanced Error Rate} (BER) are proposed in~\cite{laclau2021all}.
Specifically, DI quantifies to what level the link prediction model $f_{\text{link}}$ prefers to give positive predictions for nodes with different sensitive feature values compared with those with the same ones. It is formally formulated as 
\begin{align}
\label{di}
\text{DI} \left(f_{\text{link}}, \mathcal{G}\right)=\frac{P(f_{\text{link}} (v,v')=1|s \neq s')}{P(f_{\text{link}} (v,v')=1|s = s')}.
\end{align}
A similar strategy is also adopted by BER~\cite{laclau2021all}, which measures the difference between $P(f_{\text{link}} (v,v')=1|s = s')$ and $P(f_{\text{link}} (v,v')=1|s \neq s')$.

\noindent \textbf{Extension to Continuous Sensitive Feature(s).} Although demographic parity has been widely used, most related studies are based on categorical (either binary or multi-class) sensitive feature(s).
To extend the notion of demographic parity to continuous sensitive feature(s), Jiang et al.~\cite{jiang2021generalized} proposed \textit{Generalized Demographic Parity} (GDP) in node classification tasks. Specifically, for any sensitive feature value $s$, GDP requires that the difference between $E[\hat{Y} \mid S=s]$ and $E[\hat{Y}]$ should be as small as possible, where $E[\cdot]$ is the expectation operator.
%
%
%
We then introduce $\Delta \text{GDP}$ to quantify how well GDP is achieved:
\begin{align}
\Delta G D P =\int_{0}^{1}\left|E[\hat{Y} \mid S=s]-E[\hat{Y}]\right| f_{S}(S=s) \mathrm{d} S.
\end{align}
The value of the sensitive feature $S$ is assumed to be continuous and normalized between 0 and 1; $f_{S}(S=s)$ is the value of the sensitive feature PDF at $S=s$. 
%
Typically, a smaller value of $\Delta \text{GDP}$ indicates a higher level of $\text{GDP}$ for the corresponding algorithm.

\subsubsection{Equality of Odds}
\label{EO}

\textit{Equality of Odds} is first introduced as a group fairness notion by Hardt et al.~\cite{DBLP:conf/nips/HardtPNS16} in binary classification tasks. In general, the algorithm predictions are enforced to be independent with the sensitive feature(s) conditional on the ground truth class labels.
The rationale is to prohibit the model from abusing the sensitive feature as a proxy of class labels for prediction. We introduce equality of odds in node classification tasks as follows.

\noindent \textbf{Equality of Odds in Node Classification.} 
In node classification, suppose that the predictions of the graph mining algorithm $\hat{Y}$, the ground truth labels $Y$, and the sensitive feature $S$ are all binary.
Equality of odds requires that
\begin{align}
\label{eq_odds}
P(\hat{Y}=1 | S=0, Y=y) = P(\hat{Y}=1 | S=1, Y=y)
\end{align}
holds for both $y=0$ and $y=1$.
In other words, Eq.~(\ref{eq_odds}) enforces predictions to bear equal TPR (i.e., True Positive Rate) and FPR (i.e., False Positive Rate) for the two sensitive subgroups.
To quantify how well the equality of odds is satisfied, the largest difference of TPR (and FPR) between any two sensitive subgroups is often considered~\cite{DBLP:journals/pvldb/LahotiGW19,ma2021subgroup}.
%
%
The notion of equality of odds has also been extended to multi-class scenarios~\cite{spinelli2021biased}, where the value of TPR (and FPR) is required to be the same across all sensitive subgroups.
%
%
To compute equality of odds in multi-class scenarios, we first calculate the maximum TPR difference and the maximum FPR difference between any two subgroups. Equality of odds is then measured by the larger difference value~\cite{spinelli2021biased}.
%

\subsubsection{Equality of Opportunity}
\label{EOPP}

\textit{Equality of Opportunity} extends the notion of equality of odds~\cite{DBLP:conf/nips/HardtPNS16}. Specifically, in binary classification tasks, equality of opportunity only requires the positive predictions to be independent of sensitive feature(s) for individuals with positive ground truth labels~\cite{DBLP:conf/nips/HardtPNS16}. 
We introduce equality of opportunity in node classification tasks as follows.

\noindent \textbf{Equality of Opportunity in Node Classification.} 
The criterion of equality of opportunity is given as
\begin{align}
\label{equal_opp}
P(\hat{Y}=1 | S=0, Y=1) = P(\hat{Y}=1 | S=1, Y=1).
\end{align}
It should be noted that in most cases, $\hat{Y}=1$ is an advantaged prediction~\cite{DBLP:conf/nips/HardtPNS16}. 
Therefore, the intuition of equality of opportunity can be interpreted as: we want to avoid assigning disadvantaged predictions to individuals qualified for advantaged ones only because of their sensitive subgroup membership.
In this regard, equality of opportunity is often advocated for economic justice~\cite{binns2018fairness}, and a typical application scenario is job candidate selection.
We then introduce a commonly employed quantitative metric $\Delta_{EO}$ for equality of opportunity in node classification. Specifically, $\Delta_{EO}$ measures how far the prediction deviates from the ideal situation that satisfies equality of opportunity. $\Delta_{EO}$ is formally given as
\begin{align}
\Delta_{E O}&=|P(\hat{Y}=1 \mid Y=1, S=0) \notag \\ &-P(\hat{Y}=1 \mid Y=1, S=1)|.
\end{align}

\noindent \textbf{Extension to Link Prediction.} 
Equality of opportunity has been extended to link prediction tasks~\cite{DBLP:conf/icml/BuylB20}. At a high level, equality of opportunity in link prediction is achieved when the TPR of link prediction is independent of the underlying sensitive feature. 
Formally, the node pairs with predicted positive links are regarded as the instances with $\hat{Y}=1$, while the node pairs connected with actual links are the instances with $Y=1$.
%
%
Denote $N_{i,j}=|\{(v_l,v_m)|, v_l\in \mathcal{V}_i, v_m\in \mathcal{V}_j, \mathbf{A}_{l,m}=1\}|$, which is the number of ground truth edges between the $i$-th and the $j$-th sensitive subgroups. For node pairs spanning across the $i$-th and the $j$-th subgroups, the TPR is given as
\begin{equation}
\label{tpr-link-eo}
    \epsilon_{i,j}=\frac{1}{N_{i,j}}\sum_{v_l\in \mathcal{V}_i}\sum_{v_m\in \mathcal{V}_j}P_{v_l ,v_m}\mathbbm{1}(\mathbf{A}_{\alpha,\beta}=1).
\end{equation}
Here, $\mathbbm{1}(\cdot)$ is the identity function; $P_{v_l ,v_m}$ is the predicted probability that node $v_l$ and $v_m$ are linked.
Generally, if the TPRs of any two sensitive subgroup pairs are the same, then they are independent of the sensitive subgroup membership.
Denote ($i$, $j$) and ($k$, $l$) as the indices of two sensitive subgroup pairs (indices values can be the same within each tuple).
The criterion of equality of opportunity is then formulated as $ \epsilon_{i,j} = \epsilon_{k,l}, \,\,\forall i,j,k,l$.
%
To quantitatively measure the equality of opportunity in link prediction, 
%
%
%
Woodworth et al.~\cite{woodworth2017learning} proposed $\Delta_{EO}^{\text{link}}$, which is formally defined as
\begin{equation}
    \Delta_{EO}^{\text{link}}=\max_{\forall i,j,k,l}|\epsilon_{i,j}-\epsilon_{k,l}|.
\end{equation}

\subsubsection{Group Fairness in Node Embedding Learning}

\label{Model-based Predictive Parity}

Learning fair node embeddings has received much research attention in recent years, as these fair embeddings can be employed for various downstream tasks to achieve fair results. Nevertheless, as node embeddings are mostly leaned without using the node label information, traditional fairness notions such as demographic parity (introduced in Section~\ref{DP}), equality of odds (introduced in Section~\ref{EO}), and equality of opportunity (introduced in Section~\ref{EOPP}) cannot be directly grafted. Here we introduce two types of group fairness notions for node embeddings.

\noindent
\textbf{Distribution-Based Fairness.}
A common criterion of fulfilling fairness for node embeddings is that the distributions of learned embeddings from different sensitive subgroups are close to each other~\cite{fan2021fair,dong2021edits,navarin2020learning,du2020fairnesshete}. Generally, if the distributions of embeddings from different sensitive subgroups are similar, these embeddings are then regarded as decoupled from the sensitive feature, i.e., they are fair in terms of group fairness. 
Typically, distribution-based fairness is measured by the difference between these embedding distributions, and a common metric to quantify distribution difference is Wasserstein distance~\cite{dong2021edits,fan2021fair}.
%
%
%


\noindent
\textbf{Model-Based Fairness.}
%
The intuition of model-based fairness is to train a new model (e.g., MLP classifier~\cite{DBLP:conf/icml/BoseH19} or SVM~\cite{DBLP:conf/icml/BuylB20}) to predict the sensitive feature value based on the obtained node embeddings. 
For such a new model, the incapability of the sensitive feature prediction (e.g., low prediction accuracy) indicates the decoupling between the node embeddings and the sensitive feature, which thus implies a high level of group fairness~\cite{wu2021learning}.

\subsubsection{Group Fairness in Graph Clustering}


Group fairness in graph clustering requires all sensitive subgroups to be proportionally represented by the nodes in each cluster~\cite{DBLP:conf/icml/KleindessnerSAM19,chierichetti2018fair,schmidt2018fair}. 
In this scenario, a common fairness metric is \textit{Balance Score}~\cite{DBLP:conf/icml/KleindessnerSAM19}. 
Formally, the balance score of the cluster $\mathcal{C}_k$ ($1 \leq k \leq K$, where $K$ is the total cluster number) is given as
\begin{align}
\label{clustering_metric}
    \text{balance}(\mathcal{C}_k) = \min_{i \neq i', i,i' \in \{1,..., H\}}  \frac{|\mathcal{V}_i \cap \mathcal{C}_k|}{|\mathcal{V}_{i'} \cap \mathcal{C}_k|}, 
\end{align}
where $\mathcal{V}_i$ denotes the node set of the $i$-th sensitive subgroup; $H$ is the number of sensitive subgroups. 
Generally, the balance score of cluster $\mathcal{C}_k$ reflects the largest discrepancy of node numbers between any two sensitive subgroups in this cluster.
Considering that $\min _{k} \operatorname{balance}\left(C_{k}\right) \leq \min _{i \neq i^{\prime}}\left|V_{i}\right| /\left|V_{i^{\prime}}\right|$, a larger minimum balance score over all clusters indicates a higher level of fairness in clustering~\cite{chierichetti2018fair}.

%



\subsubsection{Group Fairness in Influence Maximization}
\label{im-fair-4}

Influence maximization algorithms have been adopted in various high-stake scenarios such as HIV prevention 
\cite{yadav2018bridging}, financial inclusion~\cite{banerjee2013diffusion}, and disease transmission~\cite{atwood2019fair}.
Recently, algorithmic fairness in influence maximization has also attracted much research attention.
In influence maximization, a set of nodes are initiated as seeds in a graph, and each seed influences its neighboring nodes by a certain probability. Given a budget for the number of seed nodes, the goal of influence maximization is to find a seed node set to influence the largest number of nodes.
Specifically, the notion of fairness in influence maximization can be defined from different perspectives as below.

\noindent \textbf{Maxmin Fairness.}
\textit{Maxmin Fairness} is first introduced by Tsang et al.~\cite{tsang2019group} based on Rawlsian theory of justice~\cite{rawls1999theory}.
%
%
Given the seed node set $\mathcal{A}$, the lowest ratio of the influenced nodes among all sensitive subgroups is given as
\begin{align}
\label{maxmin-fair}
    U_{\text{min}} =  \min_{i} \frac{\mathcal{I}_{\mathcal{G}, \mathcal{V}_{i}}(\mathcal{A})}{\left|\mathcal{V}_{i}\right|}, \text{where} \; i \in \{1, ..., H\}.
\end{align}
Here $H$ is the total number of sensitive subgroups; $\mathcal{I}_{\mathcal{G}, \mathcal{V}_{i}}(\mathcal{A})$ is the expected number of the influenced nodes in the $i$-th sensitive subgroup based on $\mathcal{A}$.
Maxmin fairness requires that $U_{\text{min}}$ should be as large as possible.
Meanwhile, the value of $U_{\text{min}}$ is employed to quantify the level of maxmin fairness in~\cite{tsang2019group}.
However, it should be noted that achieving a large $U_{\text{min}}$ can greatly jeopardize the goal of maximizing the influence over the whole population.
For instance, Tsang et al.~\cite{tsang2019group} pointed out that one sensitive subgroup can be poorly connected with other nodes in the graph. In such a case, many seed nodes would be assigned to this subgroup to promote the influence rate within such subgroup, despite the fact that they can be reassigned to other subgroups to achieve a higher influence rate over the whole population.

\noindent \textbf{Diversity.}
%
%
Tsang et al.~\cite{tsang2019group} proposed another fairness criterion named \textit{Diversity Constraint}. Here, it is assumed that a budget of seed nodes is provided for each sensitive subgroup, and the budget size is proportional to the subgroup size. 
Denote the seed node set for $\mathcal{V}_{i}$ as $\mathcal{A}_i'$, where $|\mathcal{A}_i'| = \left\lceil K\left|\mathcal{V}_{i}\right| /|\mathcal{V}|\right\rceil$. Here $K$ is the budget for the seed node set of the whole population (i.e., $\mathcal{A}$), and $\left\lceil \cdot \right\rceil$ is the ceiling function.
%
%
Then the authors defined $\mathcal{I}_{\mathcal{G}\left[\mathcal{V}_{i}\right]}\left(\mathcal{A}_i' \right)$ as the expected number of the influenced nodes in $\mathcal{V}_i$, where these nodes can only be influenced by the seed nodes in $\mathcal{A}_i'$ via intra-group edges.
Diversity constraint requires that the choice of $\mathcal{A}$ should satisfy $\mathcal{I}_{\mathcal{G}, \mathcal{V}_{i}}(\mathcal{A}) \geq \max_{\mathcal{A}_i'} \mathcal{I}_{\mathcal{G}\left[\mathcal{V}_{i}\right]}\left(\mathcal{A}_i' \right)$ for all $i$.
In other words, for each sensitive subgroup, $\mathcal{A}$ should achieve an influence rate larger or equal to the influence rate when this group is assigned a proportional number of seed nodes, given that the influence only flows via intra-group edges.
If such criterion is satisfied, the sensitive subgroups with fewer (than a proportional number) seed nodes would still obtain enough amount of influence from other sensitive subgroups.
%
%
%
The percentage of sensitive subgroups that violates such criterion is used as the metric in~\cite{tsang2019group}.

\noindent \textbf{Utility Difference-Based Fairness.} 
%
%
A different desideratum of fair influence maximization is to ensure that the influenced node ratios in different sensitive subgroups are similar.
\textit{Demographic Parity in Influence Maximization} (DP) is defined following such idea~\cite{ali2019fairness,stoica2020seeding,rahmattalabi2020fair}, and its criterion is formulated as
%
\begin{align}
\label{dpim}
 \left|u_{i}(\mathcal{A})-u_{j}(\mathcal{A})\right| \leq \delta, \,\forall i,j \in \{1,..., H\},\text { where } \delta \in [0,1).
\end{align}
%
Here $\mathcal{A}$ is the seed set within budget $K$. 
%
$\delta$ is the largest tolerance for the influenced node ratio difference between any two sensitive subgroups, while $u_i (\mathcal{A})$ is the expected ratio of influenced nodes in the $i$-th sensitive subgroup under seed set $\mathcal{A}$. 
To measure utility difference between different subgroups, \textit{Maximum Disparity in Normalized Utilities} is proposed in~\cite{ali2019fairness}. Formally, it is defined as
\begin{align}
\label{max_disp}
\max _{i, j \in\{1, \ldots, H\}}\left|u_{i}(\mathcal{A})-u_{j}(\mathcal{A})\right|.
\end{align}
Generally, Eq.~(\ref{max_disp}) measures the maximum disparity of the influenced node ratio between any two sensitive subgroups.

\noindent \textbf{Seed Set-Based Fairness.} 
In influence maximization, the selection of seed node set could also encounter unfairness issues. 
For example, if the propagation of HIV awareness is originated from a set of seed nodes with unbalanced sensitive subgroup membership, some sensitive subgroups could have the privilege to know critical knowledge about  HIV much earlier than other sensitive subgroups.
This indicates that a biased seed node set could potentially put some sensitive subgroups in disadvantaged situations~\cite{stoica2019fairness}. 
Therefore, it is critical to study the fairness of seed node selection. A common criterion of seed set-based fairness is that the number of seed nodes in any sensitive subgroup should be proportional to its population~\cite{farnad2020unifying,stoica2019fairness}. Formally, for $1 \leq i,j \leq H$ ($H$ is the total number of sensitive subgroups), the criterion is given as
\begin{align}
\label{seed}
\frac{ \left|\left\{v \in \mathcal{A} \mid v \in \mathcal{V}_{i}\right\}\right|}{\left|\mathcal{V}_{i}\right|}=\frac{\left|\left\{v \in \mathcal{A} \mid v \in \mathcal{V}_{j}\right\}\right|}{\left|\mathcal{V}_{j}\right|}.
\end{align}
%
%
%
%
To measure seed set-based fairness for sensitive subgroup $\mathcal{V}_i$, Stoica et al.~\cite{stoica2019fairness} employed the discrepancy between $|\mathcal{A} \cap \mathcal{V}_i| / |\mathcal{V}_i|$ and $|\mathcal{V}_i| / |\mathcal{V}|$.
Here a smaller discrepancy indicates a higher level of seed set-based fairness for $\mathcal{V}_i$.
%

\subsection{Individual Fairness}

\label{individual-fairness}

Compared with group fairness, \textit{Individual Fairness} does not consider any sensitive features. Instead, it focuses on fairness at the individual (e.g., each node in graph data) level~\cite{DBLP:conf/innovations/DworkHPRZ12}. Generally, individual fairness requires that similar individuals should be treated similarly. 
Hence, individual fairness is considered as a fairness notion at a finer granularity than group fairness. Currently, there are only a few graph mining algorithms that consider individual fairness. 
We hereby present several existing individual fairness notions.

\subsubsection{Node Pair Distance-Based Fairness}
\label{kang-indi}
A widely adopted definition of individual fairness is that the pair-wise node distances in the input space and output space should satisfy \textit{Lipschitz Condition}~\cite{DBLP:conf/innovations/DworkHPRZ12,DBLP:conf/kdd/KangHMT20}.
Specifically, Lipschitz condition requires that the distance of any node pairs in the output space should be smaller or equal to their corresponding distance in the input space (usually re-scaled by a scalar). Formally, given a pair of nodes $v_i$ and $v_j$, Lipschitz condition is given as
\begin{align}
D(f(v_i),f(v_j))\leq L \cdot d(v_i,v_j),
\end{align}
where $f(\cdot)$ is the predictive model that gives the node level output (e.g., node embeddings). $D(\cdot,\cdot)$ and $d(\cdot,\cdot)$ are the distance metrics in the output and input space, respectively. $L$ is the Lipschitz constant that re-scales the input distance between node $v_i$ and $v_j$.
In graph mining, Lipschitz condition formulates the guiding principle of individual fairness (i.e., treating similar input nodes similarly) by restricting the pair-wise output distance of nodes.
%

To measure individual fairness based on Lipschitz condition, Zemel et al.~\cite{DBLP:conf/icml/ZemelWSPD13} first proposed \textit{Consistency} on non-graph data.
The intuition is to measure the average distance of the output between each individual and its $k$-nearest neighbors.
Generally, for the algorithm outputs, a larger average distance indicates a lower level of individual fairness.
%
%
%
In~\cite{DBLP:conf/icml/ZemelWSPD13}, consistency is defined as 
\begin{align}
1-\frac{1}{n \cdot k} \sum_{i=1}^n\left|\hat{y}_{i}-\sum_{j \in k\text{NN}\left(\mathbf{x}_{i}\right)} \hat{y}_{j}\right|,
\end{align}
where $\hat{y}_{i}$ is the probabilistic classification output for node $v_i$; function $k\text{NN}(\cdot)$ takes the features of node $v_i$ as input and returns the index set of its $k$-nearest neighbors in the feature space.
In graph mining algorithms, a similar formulation of consistency is proposed by Lahoti et al.~\cite{DBLP:journals/pvldb/LahotiGW19} based on similarity matrix $\mathbf{S}$.
Generally, $\mathbf{S}$ describes the similarity between nodes in the input space and can be given based on attributes, graph topology, or knowledge from domain experts~\cite{DBLP:journals/pvldb/LahotiGW19,DBLP:conf/kdd/KangHMT20}. 
Here, consistency is formulated given as
\begin{align}
    1-\frac{\sum_{i} \sum_{j}\left|\hat{y}_{i}-\hat{y}_{j}\right| \cdot \mathbf{S}_{i j}}{\sum_{i} \sum_{j} \mathbf{S}_{i j}} \quad (i \neq j)
\end{align}
in binary node classification tasks. For any graph mining algorithm, a large value of consistency indicates that it gives similar outputs for similar nodes in the input space, i.e., the algorithm performs well on individual fairness.
Apart from that, Kang et al.~\cite{DBLP:conf/kdd/KangHMT20} proposed to use the similarity-weighted output discrepancy between nodes to measure unfairness, which is formulated as $\text{Tr}(\mathbf{\hat{Y}^{\top}}       \mathbf{L}_{\mathbf{S}}              \mathbf{\hat{Y}}   )$.
Here $\mathbf{\hat{Y}}$ is the output matrix of the graph mining algorithm.
Each row in $\mathbf{\hat{Y}}$ represents the output vector for the corresponding node.
%
$\text{Tr}(\cdot)$ is the trace operator for any matrix. $\mathbf{L}_{\mathbf{S}}$ denotes the Laplacian matrix of the similarity matrix $\mathbf{S}$.
Such a metric measures the weighted sum of pair-wise node distance in the output space, where the weighting score is the pair-wise node similarity.
Hence for any graph mining algorithms, a smaller value of the similarity-weighted discrepancy typically implies a higher level of individual fairness.

\subsubsection{Node Ranking-Based Fairness}

\label{individual-fairness-ranking}

%
%
%

Although Lipschitz condition (introduced in Section~\ref{kang-indi}) has been widely used as the individual fairness criterion, there could be problematic in practical scenarios. 
Specifically, determining whether the outputs of two individuals are similar or not based on absolute input distances can be inappropriate, as such criterion cannot calibrate across different individuals~\cite{dong2021individual}. 
Besides, Lipschitz condition imposes the comparison between the distances in two different spaces, which makes the Lipschitz constant hard to be determined.
%
%
To handle these drawbacks,
Dong et al.~\cite{dong2021individual} proposed to fulfill individual fairness from a ranking perspective. Specifically, a similarity matrix $\mathbf{S}$ is provided to describe the pair-wise similarity for individuals in the input space.
Based on $\mathbf{S}$, each individual has a ranking list $R_1$ that indicates the relative similarity ranking between itself and others. Similarly, a corresponding ranking list $R_2$ can also be derived based on the pair-wise individual similarity in the output space.
From the perspective of node ranking, individual fairness is regarded as fulfilled when the two ranking lists ($R_1$ and $R_2$) are the same for each individual~\cite{dong2021individual}.
Nevertheless, such a criterion is hard to be satisfied.
In practice, the average top-$k$ similarity between $R_1$ and $R_2$ over all individuals is adopted to measure individual fairness, where
NDCG@$k$~\cite{jarvelin2002cumulated} and ERR@$k$~\cite{chapelle2009expected} are two common ranking similarity metrics.
%
%


\subsubsection{Individual Fairness in Graph Clustering}
\label{sec: individual_cluster}
%
In clustering, the criterion of individual fairness is defined in a different way compared with other tasks. 
%
%
%
Specifically, for each node in a graph, if its neighbors are proportionally distributed to each cluster, individual fairness is then fulfilled~\cite{gupta2021protecting}.
Formally, under individual fairness, a clustering algorithm satisfies fair clustering for node $v_i$ if
\begin{align}
\label{indi_sc}
\frac{|\{ v_j: \mathbf{A}_{i,j}=1 \wedge v_{j} \in \mathcal{C}_k\}|}{|\mathcal{C}_k|} = \frac{|\{ v_j: \mathbf{A}_{i,j}=1\}|}{|\mathcal{V}|}
\end{align}
for all $k \in \{1, ..., K\}$.
Here $K$ represents the total number of clusters, and $\mathcal{C}_k$ represents the set of nodes in cluster $k$.
The intuition here is that for each node, the ratio occupied by its one-hop neighbors in each cluster should be the same as the ratio occupied by its one-hop neighbors in the whole population.
We then introduce the metric to quantify fair clustering.
Specifically, $\rho_i$ measures how disproportionately the one-hop neighbors of node $v_i$ are assigned in different clusters~\cite{gupta2021protecting}. 
Formally, $\rho_i$ is defined as
\begin{align}
\rho_i = \text{min}_{k,l \in \{1,..., K\}} \frac{|\mathcal{C}_k \cap \mathcal{N}_{v_i}|}{|\mathcal{C}_l \cap \mathcal{N}_{v_i}|},
\end{align}
where $\mathcal{N}_{v_i}$ denotes the neighboring node set of $v_i$. 
In~\cite{gupta2021protecting}, the average $\rho_i$ across all nodes in the graph is employed as the metric of individual fairness in graph clustering.

\subsection{Counterfactual Fairness}

Different from the above fairness notions, \textit{Counterfactual Fairness} \cite{kusner2017counterfactual} defines fairness from the causal perspective \cite{pearl2009causality}. Specifically, counterfactual fairness is considered to be achieved when the prediction results for each individual and his/her counterfactuals (``counterfactuals" in this setting are different versions of the same individual when his/her sensitive feature had been changed to different values) are maintained to be the same. For example, the algorithmic decision for an applicant's loan application should be the same regardless of his/her race. 
%
Counterfactual fairness is defined based on Pearl's structural causal model \cite{pearl2009causality}, where a causal model describes the causal relations between different variables.
For any variables $A$ and $B$, the counterfactual ``what would $B$ have been if $A$ had been set to a specific value $a$" is denoted by $B_{A\leftarrow a}$. Denote $\hat{Y}_{S\leftarrow s}=f(X_{S\leftarrow s},s)$ as the model prediction made for the counterfactual when the sensitive feature $S$ had been set to a $s$. Counterfactual fairness is defined as
\begin{equation}
P(\hat{Y}_{S\leftarrow s}=y|X=\mathbf{x},S=s) = P(\hat{Y}_{S\leftarrow s'}=y|X=\mathbf{x},S=s)
\end{equation}
for all specific values $y,\mathbf{x}$ and $s'\ne s$. 

Recently, there has been a line of works~\cite{agarwal2021towards,zhang2021multi,ma2022Learning} that extend counterfactual fairness from traditional i.i.d. data to graph data. Most of these works aim to learn counterfactually fair node embeddings, and then make predictions based on the embeddings. Agarwal et al.~\cite{agarwal2021towards} defined that a graph mining algorithm satisfies counterfactual fairness if the corresponding embedding for each node remains the same regardless of its sensitive feature(s) values (other features and graph structure stay unchanged). To step forward, Ma et al.~\cite{ma2022Learning} further considered more subtle issues that may cause counterfactual unfairness in graphs: (1) biases can be induced by each node’s neighboring nodes in graphs; (2) biases can be induced by the causal relations between the sensitive feature(s) to other features or graph structure. Accordingly, this work defines graph counterfactual fairness for node embedding learning as follows. 
Given a graph encoder $\Phi(\cdot): \mathbb{R}^{n\times d} \times \mathbb{R}^{n \times n}\rightarrow \mathbb{R}^{n\times d'}$, it satisfies graph counterfactual fairness if we have
\begin{equation}
    \Phi(\mathbf{X}_{\mathbf{S}\leftarrow \mathbf{s}'},\mathbf{A}_{\mathbf{S}\leftarrow \mathbf{s}'})_i = \Phi(\mathbf{X}_{\mathbf{S}\leftarrow \mathbf{s}''},\mathbf{A}_{\mathbf{S}\leftarrow \mathbf{s}''})_i,
\label{eq:gcf}
\end{equation}
for any node $v_i$, where $n$ is the number of nodes, $\mathbf{s}'$ and $\mathbf{s}''$ are arbitrary sensitive feature values of all nodes, where $\mathbf{s}', \mathbf{s}'' \in \{0,1\}^{n}$ and $\mathbf{s}' \ne \mathbf{s}''$,
$\Phi(\cdot)_i$ denotes the embedding of node $v_i$. Such a criterion requires the embeddings learned from the original graph and counterfactuals (where the sensitive feature values of any subset of the $n$ nodes had been changed) to be the same.

To measure counterfactual fairness on graphs, 
%
%
recent works~\cite{agarwal2021towards,zhang2021multi} usually adopt \textit{Unfairness Score}, which is the percentage of nodes whose predicted label changes when their sensitive feature values are changed (while other features are fixed). Beyond that, Ma et al.~\cite{ma2022Learning} 
also proposed to evaluate graph counterfactual fairness by measuring the average prediction discrepancy between any two different versions of counterfactual sensitive feature assignment on all the $n$ nodes. However, as there are too many combinations for the sensitive feature values of all the nodes, and the true causal model is hardly available in the real world, such metric is commonly computed by approximation \cite{ma2022Learning}.
%

\subsection{Degree-Related Fairness}
Different from other traditional fairness notions, the study on \textit{Degree-Related Fairness} is fairly new in the graph mining community.
%
%
In networked data, if two nodes are connected, there could be dependency between them. Such dependency can be informative, and thus extracting the dependency between connected nodes benefits various tasks in graph mining.
However, for low-degree nodes, their connections only contribute limited information on the dependency between themselves and other nodes. In this regard, it can be difficult for graph mining algorithms to effectively capture critical information about these nodes, which often leads to worse utility compared with high-degree nodes.
For example, the performance of GNNs in graph analytical tasks (e.g., node classification) on high-degree nodes (e.g., a celebrity who has a lot of followers) often deviates from that on low-degree nodes (e.g., an average Joe who has few followers)~\cite{kangwww2022,chen2020simple,tang2020investigating}. 
%
%
%
%
Correspondingly, degree-related fairness requires that nodes should bear similar utility (e.g., node classification accuracy) in the graph mining algorithms regardless of their degrees.
%
%
To measure degree-related fairness, Kang et al.~\cite{kangwww2022} leveraged the variance of average cross-entropy loss w.r.t. node degrees, and such metric is defined for node classification tasks.

\subsection{Application-Specific Fairness}
Apart from the application-agnostic fairness notions mentioned above, there are also other fairness notions particularly designed for certain graph-based applications. Here, we introduce application-specific fairness notions in recommender systems and knowledge graphs.

\subsubsection{Fairness Notions in Recommender Systems}
\label{modu}

\noindent \textbf{User Fairness.}
\textit{User Fairness} is an indispensable fairness notion in recommender systems. Generally, user fairness requires that the recommendation quality for different users~\cite{DBLP:conf/sigir/FuXGZHGXGSZM20,li2021user,leonhardt2018user} should be similar. For example, in a recommender system, we can divide users into two groups according to their activity levels: active users and inactive ones.
Usually, inactive users tend to receive unsatisfying recommendations compared with the active ones~\cite{DBLP:conf/sigir/FuXGZHGXGSZM20,li2021user}, as they reveal less information about their preferences. 
%
%
Correspondingly, user fairness is usually measured by the recommendation quality discrepancy between active and inactive users~\cite{li2021user}. 
Additionally, users can be divided into different sensitive subgroups according to their sensitive features. Yao et al.~\cite{DBLP:conf/nips/YaoH17} proposed to measure the disadvantage level of each sensitive subgroup with the average deviation between the predicted item ratings and ground truth ratings among group members. The average disadvantage level across all subgroups is then employed to quantify the user unfairness.
Furthermore, user fairness has also been studied in \textit{Group Recommendation}. In general, the goal of group recommendation is to provide recommendations that comply with the preferences of most users in the group~\cite{amer2009group}. However, when the preferences of certain users (in a group) are ignored by the recommender system, they would feel that they are being unfairly treated. 
User fairness then requires that the preferences of each user in a given group should not be neglected by the recommendation algorithm.
To measure user fairness for a group of users, Malecek et al.~\cite{DBLP:conf/um/MalecekP21} leveraged the difference between the minimal and maximal user-item relevance scores across all users. Similar metrics are also adopted by other works such as~\cite{DBLP:conf/recsys/KayaBT20,DBLP:conf/recsys/LinZZGLM17}.
%


\noindent \textbf{Popularity Fairness.}
\textit{Popularity Fairness}, which requires that popular items should not be over-emphasized compared with other items, is another common fairness notion in recommender systems~\cite{DBLP:conf/flairs/WasilewskiH16,DBLP:conf/sigir/ChenXLYSD20,krishnan2018adversarial}. 
A well-known example of popularity unfairness issue is the \textit{Filter Bubble}~\cite{pariser2012filter} problem, which describes the scenarios that users are isolated from less popular items or information~\cite{DBLP:conf/aaai/MasrourWYTE20,DBLP:conf/recsys/AbdollahpouriMB19}.
Typically, the level of popularity fairness is measured by the average recommendation rate of less popular instances (e.g., items, users, and social media posts)~\cite{abdollahpouri2017controlling,DBLP:conf/edbt/BorgesS20,DBLP:journals/corr/abs-2101-03584}. 
For example, in social platforms, Masrour et al.~\cite{DBLP:conf/aaai/MasrourWYTE20} extended the modularity score~\cite{newman2006modularity} on graph data to measure popularity fairness in friend recommendation.
%
%
Formally, assume that a certain partition divides users into different groups, and $M(v_i)$ and $M(v_j)$ represent the group membership for user $v_i$ and $v_j$, respectively.
To measure how less popular user groups are connected with others, the metric of popularity fairness is formulated as
\begin{align}
Q_{\text{fairness}} = \frac{1}{2|\mathcal{E}|} \sum_{i,j} ( \mathbf{A}_{i,j}  - \frac{d_{i} d_{j}}{2|\mathcal{E}|}   ) \delta(M(v_i), M(v_j)),
\end{align}
where $\delta(\cdot,\cdot)$ is the Kronecker delta function~\cite{trowbridge1998technique}; $d_{i}$ and $d_{j}$ indicate the degree of node $i$ and $j$; $|\mathcal{E}|$ is the total number of edges in the graph data. Given a friend recommendation algorithm, a lower value of $Q_{\text{fairness}}$ indicates that the algorithm yields more inter-group edges for the users in the graph. This implies that some less popular groups are encouraged to connect more with other groups, which relieves the filter bubble effect in friendship recommendation.

\noindent \textbf{Provider Fairness.}
\textit{Provider Fairness}, a.k.a. \textit{Producer Fairness}, requires that items from different providers should receive the same exposure rate to the customers. Various strategies have been proposed to measure provider fairness.
Patro et al.~\cite{patro2020fairrec} set a minimum exposure guarantee for all providers and used the number of unsatisfied providers to measure provider fairness. Liu et al.~\cite{liu2018personalizing} measured the provider diversity with the average number of providers appearing in recommendations. Boratto et al.~\cite{DBLP:journals/corr/abs-2006-04279} adopted both the user-item relevance difference and item exposure rate difference between different providers as the corresponding metrics.


\noindent \textbf{Marketing Fairness.}
\textit{Marketing Fairness} is another fairness notion in recommender systems proposed by Wan et al.~\cite{DBLP:conf/wsdm/WanNMM20}. In their paper, they pointed out that online shopping platform users are less likely to interact with items whose marketing strategy is not consistent with their identity.
For example, some gender-neutral items (e.g., armband) could be marketed using images of males~\cite{grubb1968perception,DBLP:conf/wsdm/WanNMM20}. Therefore, even if both male and female users are potential customers, female users tend to interact less with these items. In this context, male and female users are regarded as identity-consistent and identity-inconsistent users to the marketing content, respectively.
Recommender systems could then inherit such bias from data and yield biased recommendations for female (i.e., identity-inconsistent) users in the future.
%
%
To measure marketing fairness, Wan et al.~\cite{DBLP:conf/wsdm/WanNMM20} proposed to calculate the variance of recommendation errors for identity-consistent and identity-inconsistent users, and the variance discrepancy is adopted as the corresponding metric.


\subsubsection{Fairness Notions in Knowledge Graphs}

\noindent \textbf{Social Fairness.}
Knowledge graph embeddings could encode historical \textit{Social Biases}~\cite{DBLP:conf/emnlp/FisherMPC20,arduini2020adversarial} and one typical example is the stereotype that bankers are males and nurses are females~\cite{DBLP:journals/corr/abs-1912-02761}.
Such biases have been observed in different knowledge graph-based tasks including entity embedding learning~\cite{DBLP:conf/emnlp/FisherMPC20,DBLP:journals/corr/abs-1912-02761} and word embedding learning based on knowledge graphs~\cite{bolukbasi2016man,caliskan2017semantics,garg2018word}.
%
%
In recent years, various works have been proposed to measure such biases from knowledge graph embeddings to fulfill social fairness. As an example, given the embeddings of head entities representing human and sensitive relations (e.g., gender and race), Fisher et al.~\cite{DBLP:conf/emnlp/FisherMPC20} employed the prediction accuracy on tail entities (e.g., female/male when gender is the sensitive relation) as a bias metric.
%
%
%
%
%
Here a low accuracy indicates that most sensitive information has been removed from the embeddings of these head entities.
%
%
%
%
Meanwhile, social fairness in profession prediction can also be quantitatively measured through embedding perturbations~\cite{DBLP:journals/corr/abs-1912-02761}. Specifically, each human entity embedding is first perturbed to be more likely to have a certain value of the sensitive feature (e.g., gender, race, and occupation).
Such perturbed embedding is then used to predict the profession corresponding to this human entity.
The probability difference of having a certain profession between the perturbed and unperturbed embedding indicates the fairness level of the profession prediction.
Here a larger difference implies a higher level of unfairness.

%


\noindent \textbf{Path Diversity Fairness.}
Another fairness notion in knowledge graphs is \textit{Path Diversity Fairness}~\cite{DBLP:conf/sigir/FuXGZHGXGSZM20}, which is defined on \textit{Meta-Paths}. Generally, a meta-path connects different object types (nodes) with relations (edges) on knowledge graphs~\cite{sun2012mining}.
Path diversity fairness requires that the distributions of meta-paths (typically defined by counting the number of paths for different meta-paths) should be similar across different sensitive subgroups~\cite{DBLP:conf/sigir/FuXGZHGXGSZM20}.
%
%
Fu et al.~\cite{DBLP:conf/sigir/FuXGZHGXGSZM20} adopted Simpson’s Index of Diversity (SID)~\cite{simpson1949measurement} as the metric for unfairness.

\noindent \textbf{Popularity Fairness.}
\textit{Popularity Fairness} is also a critical fairness notion in knowledge graphs, and the popularity of each entity node is defined as its degree in the knowledge graph~\cite{arduini2020adversarial}. Generally, in knowledge graph completion tasks (i.e., predicting relations between entities or predicting a tail entity given head entity and a query relation), if the prediction accuracy is uniformly distributed w.r.t. entity node degrees, popularity fairness is then achieved.


\begin{figure}[!t]
    \centering
    \includegraphics[width=0.48\textwidth]{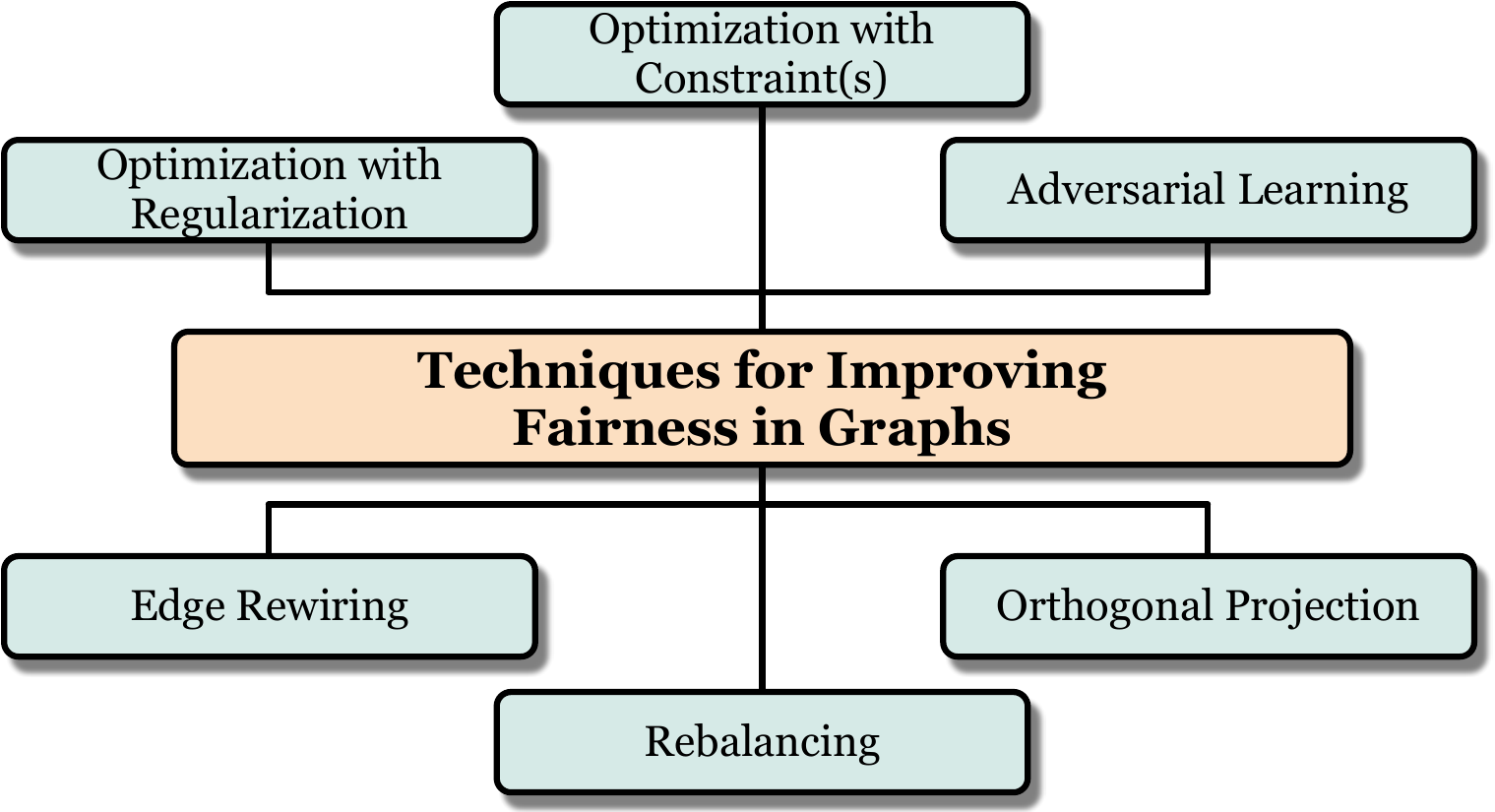}
    \vspace{-2mm}
    \caption{A taxonomy of the commonly used techniques to improve fairness in graph mining.}
    \vspace{-5mm}
    \label{tech-taxo}
\end{figure}

\begin{table*}[!h]

\caption{Surveyed publications on graph mining algorithms. Publications are categorized by the fairness notions and techniques, and all numbers of citations are collected from Google Scholar on February 20th, 2023.}
\label{summary}
\centering
\scriptsize
\renewcommand{\arraystretch}{1.05}
\setlength\tabcolsep{5.8pt}
\begin{tabular}{c|ccccc}
\hline
\textbf{Fairness Notions}                        & \textbf{Techniques}                   & \textbf{Publications}                        & \textbf{Downstream Tasks}    & \textbf{Year}     & \textbf{\# Citations}      \\  
\hline
\multicolumn{6}{l}{\textbf{Application-Agnostic}}           \\  
\hline
\multirow{41}{*}{\textbf{Group}}        
                                        & Optimization with Regularization               & \cite{kamishima2017considerations}                          & Recommendation       & 2017  & 30     \\
                                        \cline{2-6}
                                        & Optimization with Regularization                  & \cite{jiang2022fmp}                                      & Node classification  & 2022  & 9    \\ 
                                        \cline{2-6}
                                        & Optimization with Regularization               & \cite{zhang2021multi}                                       & Node classification    & 2021  & 3 \\
                                        \cline{2-6}
                                        & Optimization with Regularization               & \cite{agarwal2021towards}                    & Node classification      & 2021  & 65  \\
                                        \cline{2-6}
                                        & Optimization with Regularization               & \cite{DBLP:conf/pkdd/BuylB21}                               & Link prediction     & 2021  & 11 \\
                                        \cline{2-6}
                                        & Optimization with Regularization               & \cite{DBLP:conf/nips/YaoH17}                                & Recommendation & 2017  & 303 \\ 
                                        \cline{2-6}
                                        & Optimization with Regularization               & \cite{DBLP:journals/pvldb/LahotiGW19}                       & Node classification    & 2019  & 63        \\
                                        \cline{2-6}
                                        & Optimization with Regularization               & \cite{fan2021fair}                       & Node classification   & 2021  & 5  \\
                                        \cline{2-6}
                                        & Optimization with Regularization               & \cite{navarin2020learning}                       & Node classification   & 2020  & 4   \\
                                        \cline{2-6}
                                        & Optimization with Regularization               & \cite{franco2022deep}                       & Node classification  & 2022  & 11  \\
                                        \cline{2-6}
                                        & Optimization with Regularization               & \cite{DBLP:conf/sigir/ZhuWC20}                              & Recommendation      &2020  & 57    \\
                                        \cline{2-6}
                                        & Optimization with Regularization               & \cite{DBLP:conf/sigir/FuXGZHGXGSZM20}                       & Recommendation    &2020  & 126  \\
                                        \cline{2-6}
                                        & Optimization with Regularization               & \cite{wang2021unbiased}                       & Link prediction  &2022  & 9   \\
                                        \cline{2-6}
                                        & Optimization with Regularization               & \cite{krasanakis2020applying}                               & Recommendation         &2020  & 10    \\
                                        \cline{2-6}
                                        & Optimization with Regularization               & \cite{laclau2021all}                                        & Link prediction   &2021  & 21  \\ 
                                        \cline{2-6}
                                        & Optimization with Regularization                  & \cite{burke2017balanced}                                    & Recommendation     &2017  & 23 \\
                                        \cline{2-6}
                                        & Optimization with Constraint(s)               & \cite{DBLP:conf/icml/KleindessnerSAM19}                     & Graph clustering    &2019  & 111  \\
                                        \cline{2-6}
                                        \cline{2-6}
                                        & Optimization with Constraint(s)               & \cite{farnad2020unifying}                            & Influence maximization   &2020  & 24 \\  
                                        \cline{2-6}
                                        & Optimization with Constraint(s)               & \cite{rahmattalabi2020fair}                     & Influence maximization     &2021  & 22   \\
                                        \cline{2-6}
                                        & Optimization with Constraint(s)               & \cite{rahmattalabi2020exploring}                & Influence maximization   &2019  & 38   \\
                                        \cline{2-6}
                                        & Optimization with Constraint(s)               & \cite{ali2019fairness}                & Influence maximization    &2022  & 27  \\
                                        \cline{2-6}
                                        & Rebalancing                  & \cite{saxena2021hm}                                  & Link prediction &2021  & 6  \\
                                        \cline{2-6}
                                        & Rebalancing                  & \cite{DBLP:conf/icml/BuylB20}                               & Link prediction   &2020  & 40 \\
                                        \cline{2-6}
                                        & Rebalancing                  & \cite{zeng2021fair}                                   & Node classification     &2021  & 16  \\
                                        \cline{2-6}
                                        & Rebalancing                  & \cite{DBLP:journals/corr/abs-1809-09030}                    & Recommendation     &2018  & 39 \\
                                        \cline{2-6}
                                        & Rebalancing                  & \cite{DBLP:conf/ijcai/RahmanS0019}                          & Recommendation    &2019  & 102  \\
                                        \cline{2-6}
                                        & Rebalancing                  & \cite{tsioutsiouliklis2021fairness}                         & Node ranking &2021  & 20   \\  
                                        \cline{2-6}
                                        & Rebalancing                  & \cite{DBLP:journals/corr/abs-2105-02725}                    & Influence maximization, link prediction, and node classification  &2022  & 12   \\  
                                        \cline{2-6}
                                        & Rebalancing                  & \cite{teng2021influencing}                    & Influence maximization  &2021  & 5  \\  
                                        \cline{2-6}
                                        & Rebalancing                  & \cite{stoica2020seeding}                    & Influence maximization  &2020  & 28  \\  
                                        \cline{2-6}
                                        & Rebalancing                  & \cite{tsang2019group}                    & Influence maximization   &2019  & 75 \\  
                                        \cline{2-6}
                                        & Rebalancing                  & \cite{kose2022fair}                    & Node classification     &2022  & 8  \\
                                        \cline{2-6}
                                        & Rebalancing                  & \cite{current2022fairmod}                    & Link prediction   &2022  & 3  \\
                                        \cline{2-6}
                                        & Adversarial         & \cite{DBLP:conf/icml/BoseH19}                               & Recommendation    &2019  & 162  \\
                                        \cline{2-6}
                                        & Adversarial         & \cite{dai2021say}                                       & Node classification    &2021  & 89   \\
                                        \cline{2-6}
                                        & Adversarial         & \cite{DBLP:conf/ijcai/KhajehnejadRBHJ20}                    & Influence Maximization  &2020  & 34   \\
                                        \cline{2-6}
                                        & Adversarial         & \cite{xu2021fair}                    & Recommendation  &2021  & 1  \\
                                        \cline{2-6}
                                        & Adversarial         & \cite{wu2021learning}              & Recommendation   &2021  & 46  \\
                                        \cline{2-6}
                                        \cline{2-6}
                                        & Edge rewiring                  & \cite{li2020dyadic}                                       & Node classification  &2021  & 50 \\    
                                        \cline{2-6}
                                        & Edge rewiring                  & \cite{kose2022fair}                    & Node classification  &2022  & 8  \\ 
                                        \cline{2-6}
                                        & Edge rewiring                  & \cite{jalali2020information}                              & Topology debiasing     &2020  & 11  \\ 
                                        \cline{2-6}
                                        & Edge rewiring                  & \cite{dong2021edits}                                      & Node classification   &2022  & 28 \\ 
                                        \cline{2-6}
                                        \cline{2-6}
                                        & Edge rewiring                  & \cite{spinelli2021biased}                                 & Node classification    &2021  & 33 \\ 
                                        \cline{2-6}
                                        & Orthogonal projection        & \cite{DBLP:journals/corr/abs-1909-11793}                    & Node classification and recommendation   &2019  & 16 \\
                                        \cline{2-6}
                                        & Orthogonal projection        & \cite{palowitchdebiasing}                                   & Node classification and recommendation   &2020  & 4  \\
                                        \cline{2-6}
                                        & Orthogonal projection        & \cite{zeng2021fair}                                   & Node classification    &2021  & 16  \\
                                        \hline
\multirow{6}{*}{\textbf{Individual}}    
                                        & Optimization with Regularization               & \cite{dong2021individual}                                   & Node classification and link prediction  &2021  & 36  \\  
                                        \cline{2-6}
                                        & Optimization with Regularization               & \cite{fan2021fair}                       & Node classification   &2021  & 5   \\ 
                                        \cline{2-6}
                                        & Optimization with Regularization               & \cite{DBLP:journals/pvldb/LahotiGW19}                       & Node classification  &2019  & 63  \\
                                        \cline{2-6}
                                        & Optimization with Regularization               & \cite{DBLP:conf/kdd/KangHMT20}                              & Node ranking, node classification and graph clustering    &2020  & 61  \\
                                        \cline{2-6}
                                        & Optimization with Constraint(s)               & \cite{gupta2021protecting}                              & Graph clustering  &2021  & 6  \\
                                        \cline{2-6}
                                        & Edge rewiring               & \cite{laclau2021all}                                        & Link prediction  &2021  & 21 \\ 
                                        \hline
\multirow{3}{*}{\textbf{Degree-Related}}      
                                        & Rebalancing                  & \cite{tang2020investigating}                                & Node classification   &2020  & 53 \\
                                        \cline{2-6}
                                        & Rebalancing                  & \cite{kangwww2022}                                & Node classification  &2022  & 14  \\
                                        \cline{2-6}
                                        & Rebalancing                  & \cite{fish2019gaps}                                & Influence maximization  &2019  & 37  \\
                                        \hline
\multicolumn{6}{l}{\textbf{Application-Specific}}           \\
                                        \hline
\multirow{8}{*}{\textbf{Popularity}}    & Optimization with Regularization               & \cite{DBLP:conf/recsys/KamishimaAAS13}                      & Recommendation       &2013  & 33 \\
                                        \cline{2-6}
                                        & Optimization with Regularization               & \cite{DBLP:conf/flairs/WasilewskiH16}                       & Recommendation   &2016  & 36 \\
                                        \cline{2-6}
                                        & Optimization with Regularization               & \cite{abdollahpouri2017controlling}                   & Recommendation     &2017  & 282  \\
                                        \cline{2-6}
                                        & Optimization with Regularization               & \cite{DBLP:conf/sigir/ChenXLYSD20}                          & Recommendation  &2020  & 53 \\
                                        \cline{2-6}
                                        & Optimization with Regularization               & \cite{zhu2021popularity}                                    & Recommendation  &2021  & 48  \\
                                        \cline{2-6}
                                        & Optimization with Regularization               & \cite{DBLP:journals/corr/abs-2101-03584}                    & Recommendation   &2021  & 102   \\
                                        \cline{2-6}
                                        & Adversarial         & \cite{krishnan2018adversarial}                         & Recommendation   &2018  & 43 \\
                                        \cline{2-6}
                                        & Edge rewiring + Adversarial                  & \cite{DBLP:conf/aaai/MasrourWYTE20}                        & Link prediction &2020  & 50  \\
                                        \hline
\multirow{3}{*}{\textbf{Provider}}      
                                        & Optimization with Regularization               & \cite{liu2018personalizing}                    & Recommendation   &2018  & 45  \\
                                        \cline{2-6}
                                        & Rebalancing & \cite{DBLP:journals/corr/abs-2006-04279}                    & Recommendation &2021  & 26  \\ 
                                        \cline{2-6}
                                        & Rebalancing                  & \cite{patro2020fairrec}                            & Recommendation  &2020  & 141 \\
                                        \hline
\multirow{4}{*}{\textbf{User}}   
                                   & Optimization with Regularization               & \cite{DBLP:conf/recsys/LinZZGLM17}                         & Recommendation  &2017  & 145 \\
                                   \cline{2-6}
                                  & Optimization with Regularization               & \cite{li2021user}                                             & Recommendation  &2021  & 79  \\ 
                                  \cline{2-6}
                                   & Rebalancing               & \cite{DBLP:conf/recsys/KayaBT20}                              & Recommendation    &2020  & 34  \\
                                   \cline{2-6}
                                  & Rebalancing               & \cite{DBLP:conf/um/MalecekP21}                                & Recommendation &2021  & 8 \\ 
\hline
\textbf{Marketing}                      & Optimization with Regularization               & \cite{DBLP:conf/wsdm/WanNMM20}                              & Recommendation   &2020  & 35   \\
\hline
\multirow{3}{*}{\textbf{Social}}                      & Optimization with Regularization               & \cite{DBLP:journals/corr/abs-1912-02761}                              & Knowledge graph embedding learning  &2019  & 33  \\
                                     \cline{2-6}
                                     & Adversarial               & \cite{DBLP:conf/emnlp/FisherMPC20}                              & Triple prediction    &2020  & 24  \\
                                     \cline{2-6}
                                     & Adversarial         & \cite{arduini2020adversarial}                    & Link prediction   &2020  & 22 \\
                                        \hline
\end{tabular}
\end{table*}

\section{Techniques for Improving Fairness}
\label{technique_section}

In this section, we introduce existing techniques for improving fairness in graph mining algorithms. Generally, these techniques can be divided into six categories, namely optimization with regularization, optimization with constraint(s), rebalancing, adversarial learning, edge rewiring, and orthogonal projection. We present the taxonomy of techniques in Fig.~\ref{tech-taxo}. 
%
%
For the techniques in each category, we introduce how they promote application-agnostic and application-specific (if applicable) fairness.
The surveyed literature is summarized in Table~\ref{summary}.
Furthermore, we provide a comparison of different fairness-improving techniques under the same fairness notions in Appendix~\ref{comparison_under_same_notion}.

\begin{figure}[!t]
    \centering
    \includegraphics[width=0.48\textwidth]{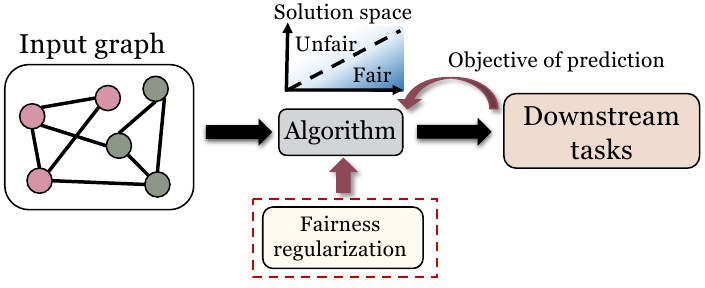}
    \vspace{-3mm}
    \caption{The pipeline of optimization with regularization. The fairness regularization encourages the optimization result to stay in the area with a higher level of fairness (i.e., more blue) of the solution space.}
    \vspace{-5mm}
    \label{reg}
\end{figure}

\subsection{Optimization with Regularization}
Optimization with regularization is a straightforward strategy to relieve unfairness in graph mining algorithms, and thus it is widely used among existing works. 
In general, the idea of regularization is to add an extra term to the optimization objective to promote the fairness level of the algorithm output. Formally, the total objective $\mathscr{L}$ is given as
\begin{align}
\mathscr{L} = \mathscr{L}_{\text{utility}} + \lambda \mathscr{L}_{\text{fair}},
\end{align}
where $\mathscr{L}_{\text{utility}}$ and $\mathscr{L}_{\text{fair}}$ are the objectives of utility and fairness, respectively; $\lambda$ controls the effect of the fairness regularization.
Compared with debiasing via regularization in the context of i.i.d. data~\cite{goel2018non,huang2019stable}, the regularization in graph mining can also be developed based on the input relational information.
We present a basic pipeline in Fig.~\ref{reg}.

\subsubsection{Improving Group Fairness}

\noindent \textbf{Algorithm Output-Based Regularization.}
%
%
Generally, the regularization enforcing statistical parity or equal opportunity is usually defined based on the algorithm output. 
For example, when the sensitive feature is binary, Zeng et al.~\cite{zeng2021fair} defined the regularization terms to enforce statistical parity and equal opportunity in node classification tasks. As an example, the regularization for statistical parity is
\begin{align}
\mathscr{L}_{s p}=\sum_{j=1}^{c} &\left(\frac{\sum\limits_{v_i \in \mathcal{V}_0} P(\hat{Y}=j \mid v_i)}{|\mathcal{V}_0|}-\frac{\sum\limits_{v_i \in \mathcal{V}_1} P(\hat{Y}=j \mid v_i)}{|\mathcal{V}_1|}\right)^{2}.
\end{align}
Here, $\mathcal{V}_0$ and $\mathcal{V}_1$ are the node sets for the two sensitive subgroups ($S=0$ and $S=1$), respectively; $\mathcal{V}_{0}^j$ and $\mathcal{V}_{1}^j$ are the sets of nodes that belong to class $j$ in the two sensitive subgroups, respectively.
%
%
Other recent works also follow a similar idea to design their regularizations for group fairness~\cite{navarin2020learning,DBLP:conf/nips/YaoH17,wang2021unbiased,franco2022deep}.

%
%

\noindent \textbf{Network Topology-Based Regularization.}
Regularization based on the network topology is proved to be effective in promoting group fairness. 
For instance, feature propagation is a common operation to model the dependency between neighboring nodes in graph mining~\cite{jiang2022fmp,shen2018mining}. However, if the graph topology is biased, the propagated features also tend to be biased~\cite{dong2021edits}.
To tackle such problem, Jiang et al.~\cite{jiang2022fmp} achieved a less biased feature propagation through a fairness-aware regularization. Specifically, given two sensitive subgroups, the regularization is formally given as
\begin{align}
\label{reg_fair}
\mathscr{L}_{\text{fair}} = \| \bm{\Delta}_s \text{softmax}(\mathbf{\hat{X}})\|_1.
\end{align}
Here $\text{softmax}(\cdot)$ is the softmax function; $\mathbf{\hat{X}} \in \mathbb{R}^{n \times d}$ is the node feature matrix after propagation; $\bm{\Delta}_s \in \{1, -1\}^{1 \times n}$ is the element-wise indicator for the sensitive subgroup membership, which is formulated as
\begin{align}
\bm{\Delta}_s = \frac{\mathbbm{1}_{=1}(\mathbf{s})}{\|\mathbbm{1}_{=1}(\mathbf{s})\|_{1}}-\frac{\mathbbm{1}_{=0}(\mathbf{s})}{\|\mathbbm{1}_{=0}(\mathbf{s})\|_{1}},
\end{align}
where $\bm{s} \in \{0,1\}^{n}$ is the sensitive feature vector for all nodes; $\mathbbm{1}_{=0}(\cdot)$ and $\mathbbm{1}_{=1}(\cdot)$ is the element-wise indicator function for 0 and 1 entries, respectively. 
As such, $\bm{\Delta}_s$ is a vector where entries corresponding to members in the two sensitive subgroups are $1$ and $-1$ normalized by their subgroup size, respectively.
This regularization encourages the average feature values after propagation to be similar between the two subgroups at each dimension.
Additionally, regularization based on the network topology is also widely employed in link prediction.
%
%
%
%
As an example, Buyl et al.~\cite{DBLP:conf/pkdd/BuylB21} proposed a fairness regularization term for the link prediction task. They first defined a set of probabilistic graph models that are fair w.r.t. demographic parity and equal opportunity. Then they measured the KL-divergence between the predicted edge distribution (by the link prediction model) and the edge distribution determined by its closest graph model. Such KL-divergence is employed as the regularization term, and minimizing it encourages the predicted edge distribution to be close to a pre-defined fair edge distribution.

\noindent \textbf{Node Embedding-Based Regularization.}
Regularization based on node embeddings is another common approach to improve group fairness. 
For example, Lahoti et al.~\cite{DBLP:journals/pvldb/LahotiGW19} utilized the total Euclidean distance of all embedding pairs spanning across different sensitive subgroups as a regularization term, which encourages the node embeddings in different sensitive subgroups to be similar. 
Apart from that, in binary sensitive feature scenarios, the distribution distance of node embeddings between the two sensitive subgroups is also an effective regularization term that helps to fulfill group fairness~\cite{navarin2020learning,fan2021fair}.


\subsubsection{Improving Individual Fairness}

\noindent \textbf{Algorithm Output-Based Regularization.}
Regularization can also be adopted to improve individual fairness for graph mining algorithms. 
A basic desideratum here is to employ regularization terms to reduce the output difference between nodes that are similar, which is consistent with the intuition ``to treat similar individuals similarly".
%
%
For example, Kang et al.~\cite{DBLP:conf/kdd/KangHMT20} leveraged an oracle similarity matrix $\mathbf{S}$ to indicate the similarity between individuals,  
%
%
and the Laplacian matrix $\mathbf{L}_{\mathbf{S}}$ for $\mathbf{S}$ can then be derived. 
The total variation of the output matrix $\mathbf{\hat{Y}}$ w.r.t. $\mathbf{S}$ is used as the regularization term, which can be formulated as $\text{Tr} (\mathbf{\hat{Y}}^{\top} \mathbf{L}_{\mathbf{S}} \mathbf{\hat{Y}})$. With such a regularization, the algorithm yields similar outputs for similar nodes, which aligns with the definition of individual fairness in Section \ref{kang-indi}. Lahoti et al.~\cite{DBLP:journals/pvldb/LahotiGW19} followed a similar idea. In their work, the similarity between nodes is derived from both node features and human knowledge.
%
%
As another example, Dong et al.~\cite{dong2021individual} formulated a regularization term based on node rankings to promote individual fairness in GNNs. Specifically, for each individual, a ranking list is first derived based on the input similarity scores between itself and other individuals. Similarly, another ranking list can also be derived based on the GNN output similarity between this individual and others. Such a regularization encourages the two ranking lists for each individual to be as similar as possible, which promotes the node ranking-based fairness.
%


\noindent \textbf{Node Embedding-Based Regularization.}
Promoting the level of group fairness based on node embedding distributions also helps to impose individual fairness~\cite{DBLP:conf/innovations/DworkHPRZ12,fan2021fair}.
Fan et al.~\cite{fan2021fair} empirically proved that the Wasserstein distance between node embedding distributions across different sensitive subgroups can be utilized as an effective regularization to improve fairness at both group and individual level in node classification tasks.

\subsubsection{Improving Counterfactual Fairness}


The intuition of counterfactual fairness on graphs is that the prediction of each individual (node) should be the same on the factual data and counterfactuals (in counterfactuals, the values of nodes' sensitive feature have been changed). Based on such intuition, many studies have been devoted in recent years. Among them, Agarwal et al. proposed NIFTY \cite{agarwal2021towards}, which generates the graph counterfactuals by flipping the sensitive feature values for all nodes while keeping everything else unchanged. A regularization in the loss function is then introduced 
to encourage the node embeddings learned from the factual graph and its counterfactual to be the same. At a high level, the regularization for counterfactual fairness can be expressed as
$
    E [D(\mathbf{z}_i, \mathbf{z}'_i)],
$
where $\mathbf{z}_i$ and $\mathbf{z}'_i$ are embeddings of node $v_i$ learned based on the factual graph and its counterfactual, respectively; $D(\cdot,\cdot)$ is a distance metric. 
%
%
A following work~\cite{ma2022Learning} proposed a similar regularization term. The main difference is that more causal relations are considered to generate graph counterfactuals. 

\subsubsection{Regularization in Applications}

\noindent \textbf{Recommender Systems.} In recommender systems, regularization is a commonly used technique to fulfill popularity fairness.
As an example, in online shopping platforms, the number of feedback actions an item receives generally represents how popular this item is. Based on such intuition, Zhu et al.~\cite{zhu2021popularity} proposed to formulate a regularization as
\begin{align}
\mathscr{L}_{fair} = \text{Corr}_{\text{P}}(\hat{\mathbf{r}}_{+}, \mathbf{p}_{+}).
\end{align}
Here $\hat{\mathbf{r}}_{+}$ denotes the vector of predicted relevance scores for positive user-item pairs; $\mathbf{p}_{+}$ represents the vector of the feedback number received by the corresponding items in user-item pairs;
$\text{Corr}_{\text{P}}(\cdot, \cdot)$ is the Pearson correlation function.
By regularizing the correlation between $\hat{\mathbf{r}}_{+}$ and $\mathbf{p}_{+}$, the effect that popular items tend to receive higher relevance scores can be relieved.
Various other works also employed regularization terms based on the popularity of items to achieve popularity fairness~\cite{DBLP:conf/recsys/KamishimaAAS13,DBLP:journals/corr/abs-2101-03584,abdollahpouri2017controlling,DBLP:conf/flairs/WasilewskiH16,DBLP:conf/sigir/ChenXLYSD20}.
Regularization is also a widely used technique to promote user fairness in recommender systems. For example, Lin et al.~\cite{DBLP:conf/recsys/LinZZGLM17} pointed out that all users should receive recommendations with equal quality. Correspondingly, the difference of recommendation quality (measured by the relevance scores of recommended items to users) between every two individuals is summed up and formulated as the regularization term. Besides, Li et al.~\cite{li2021user} defined a regularization term by considering the recommendation quality discrepancy between active users and inactive ones.
Furthermore, regularization can also be adopted to promote other types of application-specific fairness.
For example, the provider diversity of recommended items is often employed as a regularization~\cite{DBLP:journals/corr/abs-2006-04279,liu2018personalizing}. To fulfill marketing fairness, the variances of errors between recommendations to identity-consistent and identity-inconsistent users (see definitions in Section~\ref{modu}) are first computed, and their discrepancy is regarded as a regularization~\cite{DBLP:conf/wsdm/WanNMM20}.



\noindent \textbf{Knowledge Graphs.} Regularization is usually formulated in knowledge graph embedding learning to fulfill fairness for human entities. For example, Fisher~\cite{DBLP:journals/corr/abs-1912-02761} proposed to relieve social bias through regularization. 
Specifically, a prediction on the sensitive feature value is made for each human entity based on the learned embeddings. The KL-divergence between the predicted value distribution and uniform distribution over all possible sensitive feature values is then defined as the regularization term. Generally, a smaller KL-divergence indicates that the embeddings provide less information about the sensitive feature, which implies a higher level of fairness.


\noindent \textbf{Other Applications.} Regularization has also been exploited to fulfill fairness in other application scenarios. For example, Agarwal et al.~\cite{agarwal2021towards} adopted regularization towards fair node classification in criminal justice. Specifically, they aim to predict whether a defendant deserves bail over a similarity network between defendants. Moreover, fairness in default and credit risk prediction over the network between bank clients are also explored~\cite{agarwal2021towards,zhang2021multi}.

\subsection{Optimization with Constraint(s)}

\begin{figure}[!t]
    \centering
    \includegraphics[width=0.47\textwidth]{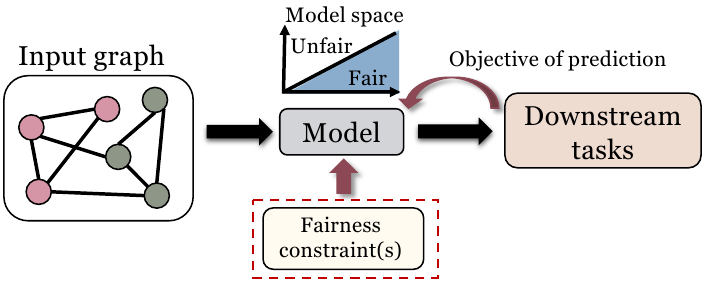}
    \vspace{-2mm}
    \caption{The pipeline of optimization with constraint(s). The formulated hard constraint for fairness restricts the solution to be in the fair area (i.e., blue area) of the solution space.}
    \vspace{-6mm}
    \label{const}
\end{figure}

Optimization with constraint(s) serves as a critical technique to fulfill fairness in graph mining.
Generally, such fairness constraint(s) reduce the feasible set size of the corresponding optimization problem to exclude unfair solutions.
%
%
We present a conceptual formulation of optimization problems with fairness constraint(s) as:
%
\begin{align}
\min& \;\; &&\mathscr{L}_{\text{utility}}, \notag \\
\text{subject}\;\; \text{to}& \;\; && \text{certain fairness constraint(s)}
\end{align}
%
%
We present a pipeline of optimization with constraint(s) in Fig.~\ref{const}. Different from regularization, optimization with constraint(s) requires that the solution should only be in the fair area (i.e., the blue area) of the solution space.
Compared with the fairness constraints based on i.i.d. data~\cite{haas2019price,kim2018fairness}, constraints in graph mining are usually tailored to specific algorithmic output patterns, e.g., seeding choices among demographic subgroups (influence maximization task).

\subsubsection{Improving Group Fairness}



Optimization with constraint(s) is widely leveraged in influence maximization to ensure the influence propagates to different sensitive subgroups in a fair manner~\cite{rahmattalabi2020exploring,ali2019fairness,farnad2020unifying}.
%
%
Generally, the fairness-constrained influence maximization problem is formulated as
\begin{align}
\label{fair_infor_max}
\begin{array}{l}\max  \underbrace{\sum_{i=1}^{H} u_{i}(\mathcal{A})}_{\text {Expected number of influenced nodes }} \\ \text { subject to } \underbrace{|\mathcal{A}| \leq K}_{\text {Bound of seed set size }} \\ \text { and } \underbrace{ M(\mathcal{A}, u_1, ..., u_H, \mathcal{V}_1, ..., \mathcal{V}_H) \leq \beta_{\text{unfair}}}_{\text {Fairness constraint}}\end{array}
\end{align}
Here $u_{i}(\mathcal{A})$ is the utility (i.e., the percentage of influenced nodes) of the $i$-th sensitive subgroup based on the seed node set $\mathcal{A}$; $K$ represents the budget for seed set size; function $M$ outputs the unfairness level of the influence maximization algorithm according to certain unfairness metric; $\beta_{\text{unfair}}$ denotes the maximum acceptable threshold for the influence maximization unfairness.
%
%
It is worth noting that the constrained optimization problem given by Eq.~(\ref{fair_infor_max}) has been proved to be an NP-hard problem~\cite{ali2019fairness}. Hence the optimization problem introduced above is usually transformed into its surrogate problem and solved in a heuristic manner.
%

Aside from influence maximization, optimization with constraint(s) is also used to fulfill fairness in other graph mining scenarios such as graph clustering.
%
As an example, Matth{\"{a}}us et al.~\cite{DBLP:conf/icml/KleindessnerSAM19} defined a fairness-aware constraint for spectral clustering to ensure that each sensitive subgroup is proportionally represented by each cluster. 
Formally, assume nodes in the input graph are divided into $H$ sensitive subgroups, and $\mathcal{V}_i$ is the node set of the $i$-th subgroup. $K$ and $\mathcal{C}_k$ are the cluster number and the node set of the $k$-th cluster, respectively. The constraint is given as 
\begin{align}
\forall i \in \{1,..., H\} \; \text{and} \; \forall k \in &\{1,..., K\}, \frac{|\mathcal{V}_i \cap \mathcal{C}_k|}{|\mathcal{C}_k|} = \frac{|\mathcal{V}_i|}{|\mathcal{V}|}.
\end{align}

\subsubsection{Improving Individual Fairness}


Optimization with constraint(s) is also a popular technique to fulfill individual fairness.
For example, Gupta et al.~\cite{gupta2021protecting} proposed to achieve individual fairness in graph clustering via optimization constraints. 
As introduced in Section \ref{sec: individual_cluster}, individual fairness in graph clustering requires that for each node in the graph, its neighbors should be proportionally assigned to different clusters. To this end, 
the corresponding constraint is formulated as 
$\forall k\in \{1, ... , K\}, \frac{1}{\left|\mathcal{C}_{k}\right|} \left|\left\{v_{j}: \mathbf{A}_{i j}=1 \wedge v_{j} \in \mathcal{C}_{k}\right\}\right| =\frac{1}{|\mathcal{V}|} \left|\left\{v_{j}: \mathbf{A}_{i j}=1\right\}\right|$ for node $v_i$, where $1 \leq i \leq n$.
To be more concise, such constraint can be further formulated as
\begin{align}
\label{sc_fair_condition}
\mathbf{A} (\mathbf{I} - \mathbf{1} \mathbf{1}^{\top}/n ) \mathbf{H} = \mathbf{0},
\end{align}
where $\mathbf{1} \in \{1\}^{n \times 1}$ and $\mathbf{0} \in \{0\}^{n \times K}$. For $k \in \{1,...,K\}$, matrix $\mathbf{H}$ is defined as
\begin{align}
\label{sc_h}
\mathbf{H}_{i,k}= 
\left\{
\begin{array}{cl}
1 / \sqrt{|\mathcal{C}_k|}, &\mbox{if}\; v_i \in \mathcal{C}_k, \\
\;0,  &\mbox{otherwise}.
\end{array}
\right.  
\end{align}

\subsection{Rebalancing}  
\begin{figure}[!t]
    \centering
    \includegraphics[width=0.48\textwidth]{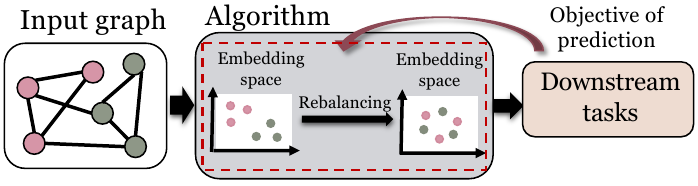}
    \vspace{-2mm}
    \caption{The pipeline of rebalancing-based approaches. Generally, rebalancing aims to make certain characteristics to be as balanced as possible between the advantaged and disadvantaged individuals.}
    \vspace{-5mm}
    \label{rebal}
\end{figure} 
Rebalancing aims to reduce the distribution difference of certain properties (e.g., the appearance rate of a node in random walks and frequency of recommendations of an item) between advantaged and disadvantaged nodes. 
We present a basic pipeline in Fig.~\ref{rebal}.
Compared with the fairness constraints in the context of i.i.d. data~\cite{krasanakis2018adaptive,jiang2020identifying}, rebalancing methods in graph mining are mostly designed based on how the algorithm utilizes the topological information.

\subsubsection{Improving Group Fairness}   

\noindent \textbf{Edge/Path-Based Rebalancing.}
A number of works adopt the rebalancing strategy to promote group fairness based on the edges or paths in the input network data.
As an example, Rahman et al.~\cite{DBLP:conf/ijcai/RahmanS0019} proposed Fairwalk to achieve fair node embedding learning, where the node appearance frequency in random walks (generated by \textit{node2vec}~\cite{grover2016node2vec}) is balanced between minority and majority groups.
%
However, when a walk is choosing its next step in FairWalk, only one-hop neighbors of the node at the current step are considered.
Such a strategy lacks long-term rebalancing and thus could fail when all nodes up to several hops away from the current step have the same membership (i.e., the current step has a homogeneous surrounding).
To consider long-term rebalancing, Khajehnejad et al.~\cite{DBLP:journals/corr/abs-2105-02725} proposed CrossWalk, which extends the rebalancing range to the whole walk by assigning larger transition probabilities to nodes that are closer to the sensitive groups’ topological peripheries.
Compared with FairWalk, CrossWalk further promotes the diversity of node membership by encouraging walks to avoid being stuck in homogeneous surroundings.
%
Moreover, in link prediction tasks, Saxena et al.~\cite{saxena2021hm} rebalanced the number of intra-group and inter-group links, which also effectively enforces group fairness.
In addition, on heterogeneous graphs, path-based rebalancing can be adopted for fair embedding learning. For example, during meta-path generation, the probabilities of selecting nodes from different sensitive subgroups are rebalanced in~\cite{zeng2021fair}.
%
Specifically, rebalancing is achieved via selecting nodes in disadvantaged subgroups with a higher probability. Such a strategy enforces a fair appearance rate for nodes from different subgroups in the generated meta-paths.

\noindent \textbf{Node Sampling/Generation-Based Rebalancing.}
Rebalancing can also be achieved via node sampling or node generation. Generally, both approaches can rebalance the node number between different sensitive subgroups. For node sampling, 
Kose et al.~\cite{kose2022fair} pointed out that if a GNN model is trained on a sampled subgraph with balanced populations from different sensitive subgroups, its predictions tend to be with a higher level of group fairness.
Another example is the fairness-aware PageRank~\cite{tsioutsiouliklis2021fairness}.
Generally, the node importance vector given by PageRank~\cite{brin1998anatomy} is derived based on the transition matrix and the jump vector. However, they could be biased due to the imbalanced size of different sensitive subgroups. By rebalancing the transition (jumping) probabilities across different sensitive subgroups, group fairness can be achieved in the fairness-aware PageRank algorithm.
For node generation, Current et al.~\cite{current2022fairmod} proposed to generate pseudo nodes and reweight edges for the input network data of GNNs to encourage a balanced information propagation in different sensitive subgroups. In their paper, such modifications on the input network are jointly optimized with the GNN model parameters.

\noindent \textbf{Information Flow-Based Rebalancing.}
Information flow-based rebalancing techniques are commonly adopted to achieve fair influence maximization. 
For example, Stoica et al.~\cite{stoica2020seeding} proposed \textit{Parity Seeding}, which is achieved by setting different seed number budgets for different sensitive subgroups. In this way, the flow of influence originating from seed nodes is rebalanced across different subgroups.
Besides, Tsang et al.~\cite{tsang2019group} rebalanced the selection of seed nodes to improve the lowest ratio of influenced nodes among all sensitive subgroups.


%

\subsubsection{Improving Degree-Related Fairness}
Degree-related fairness can also be achieved via rebalancing techniques. For example, in the message-passing process of traditional GNNs, nodes with low degrees usually benefit less (compared with nodes with high degrees) from the information propagation due to their sparse connections~\cite{kangwww2022}.
Tang et al.~\cite{tang2020investigating} proposed to rebalance the labeled nodes across the graph.
Specifically, pseudo labels are generated to improve the probability of labeled nodes appearing in the neighborhood of low-degree nodes.
In this way, more supervision information can be accessed by those low-degree nodes through the given network topology. Such a rebalancing strategy has been proved to be effective in improving the node classification accuracy for low-degree nodes.
Besides, Kang et al.~\cite{kangwww2022} pointed out that a critical source of degree-related unfairness in GNNs is the gradient of learnable weight parameters w.r.t. the objective function.
In particular, it has been proved that high-degree nodes tend to exert a more significant influence on the gradient of the learnable weight matrix, which is the reason why GNNs favor high-degree nodes.
To handle this problem, a doubly stochastic adjacency matrix (the rows and columns sum up to 1) of the GNN input network is defined and employed as GNN input. Such a strategy rebalances the influence of each node to the learnable weight matrix during optimization, which helps to enforce degree-related fairness.

Additionally, in influence maximization, Fish et al.~\cite{fish2019gaps} designed a \textit{Social Welfare Function} to measure the difficulty for the low-degree nodes to get access to the information originating from the seed nodes. Based on the welfare function, the probability that the seed nodes reach out and influence those low-degree nodes is promoted. Such a strategy helps to enable nodes with high and low degrees to receive a more balanced amount of influence.




\subsubsection{Rebalancing in Applications}


\noindent \textbf{Recommender Systems.} Upsampling is a common rebalancing approach in recommender systems. For example, in terms of provider fairness, Boratto et al.~\cite{DBLP:journals/corr/abs-2006-04279} proposed to upsample interactions between users and items from minority providers. A similar rebalancing idea is also adopted by Gourab et al.~\cite{patro2020fairrec}, where copies of products are made to improve the exposure of items from minority providers.
Rebalancing can also be leveraged to improve user fairness. For example, in group recommendation, items are recommended to a group of users.
To ensure that the preference of each user is proportionally represented by the recommended items, 
Malecek et al.~\cite{DBLP:conf/um/MalecekP21} proposed to enforce a cap on the recommendation relevance score summation for each user to rebalance the recommendation quality. 
%
%
In this way, the phenomenon that some users in a group may be under-represented and receive unsatisfying recommendation results can be eliminated.
Similar rebalancing approaches have also been applied to rebalance item ratings given by users from different sensitive subgroups to achieve a higher level of user fairness~\cite{DBLP:journals/corr/abs-1809-09030}.

\noindent \textbf{Other Applications.} Rebalancing has been adopted to fulfill fairness in real-world applications other than recommender systems. For example, Teng et al.~\cite{teng2021influencing} utilized rebalancing to realize fair information diffusion over social networks; Tsang et al.~\cite{tsang2019group} used rebalancing to prevent homeless youth from HIV over real-world social connections.

\subsection{Adversarial Learning}
\begin{figure}[!t]
    \centering
    \includegraphics[width=0.48\textwidth]{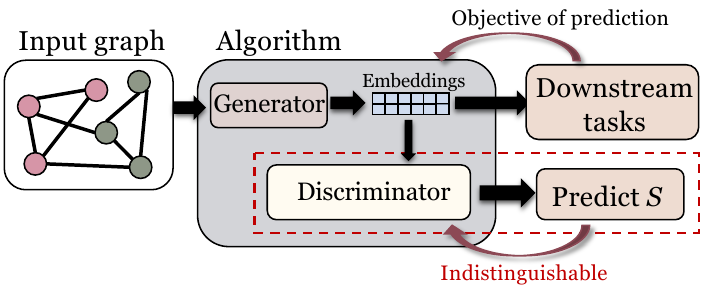}
    \vspace{-2mm}
    \caption{The pipeline of adversarial learning-based approaches. The information of sensitive feature(s) is removed from the embeddings if the discriminator cannot accurately predict sensitive feature values.}
    \vspace{-5mm}
    \label{adv}
\end{figure} 

Adversarial learning-based approaches in graph mining are naturally generalized from similar approaches in the context of i.i.d. data, e.g., ~\cite{celis2019improved,edwards2015censoring}.
In general, an adversarial learning-based framework includes a generator and a discriminator. The generator outputs node embeddings or probabilistic predictions, while the discriminator aims to predict the sensitive feature values based on the generator output.
The basic rationale here is to play a min-max game between the generator and discriminator. When the discriminator fails to predict sensitive feature values, the generator output is regarded as decoupled from the sensitive feature(s)~\cite{DBLP:conf/icml/BoseH19}. We present a basic pipeline of adversarial learning-based approaches in Fig.~\ref{adv}.

\subsubsection{Improving Group Fairness}  
Adversarial learning is a popular strategy for learning node embeddings that are fair in terms of group fairness. 
For example, Bose et al.~\cite{DBLP:conf/icml/BoseH19} leveraged a discriminator to predict the value of sensitive feature(s) based on learned node embeddings, while the generator aims to generate embeddings that are indistinguishable w.r.t. sensitive feature(s).
This idea is also followed by many other works~\cite{wu2021learning,liu2022dual,xu2021fair,dai2021say} to filter out the information of sensitive features from the learned node embeddings.
%
%
%
%
In~\cite{DBLP:conf/ijcai/KhajehnejadRBHJ20}, Khajehnejad et al. proposed to learn node embeddings based on adversarial learning to promote group fairness for influence maximization. 
%
%
With the learned embeddings, seed nodes are selected based on embedding clustering: the nodes nearest to the centroid of each cluster are selected. Considering that the information of sensitive features has been removed from the learned embeddings, the seed selection is regarded as fair.


\subsubsection{Adversarial Learning in Applications}

\noindent \textbf{Recommender Systems.} Adversarial learning has become popular in achieving fair recommendations over the years. For example, Wu et al.~\cite{wu2021fairness} proposed to utilize adversarial learning to avoid delivering news with biased content towards certain demographic subgroups. Fair recommendations are also achieved via adversarial learning in multiple recent works~\cite{liu2022dual,liu2022mitigating}.

\noindent \textbf{Knowledge Graphs.} Adversarial learning can also be adopted to improve social fairness in knowledge graph embedding learning. 
Arduini et al.~\cite{arduini2020adversarial} proposed to leverage a sensitive information filter to remove social bias from the embeddings of human entities. Here the filter plays a min-max game with a discriminator, such that the entity embeddings are decoupled from the sensitive feature(s).



\subsection{Edge Rewiring}
\begin{figure}[!t]
    \centering
    \includegraphics[width=0.48\textwidth]{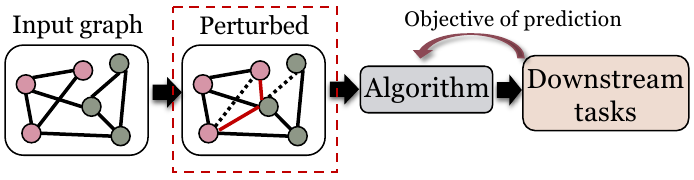}
    \vspace{-2mm}
    \caption{The pipeline of edge rewiring-based approaches. Edges in the input graph data is rewired (i.e., removed from linked node pairs and added to non-linked node pairs) to obtain fair algorithm output in downstream tasks.}
    \vspace{-5mm}
    \label{rewire}
\end{figure} 
Biases exhibited in the node embeddings and algorithm predictions could also be attributed to the biased network topology. In this regard, modifying the graph topology through edge rewiring is a common debiasing strategy.
We present a basic pipeline in Fig.~\ref{rewire}. 
It is worth noting that edge rewiring is a unique debiasing technique in graph mining compared with those algorithms centered on i.i.d. data.
%



\subsubsection{Improving Group Fairness}

\noindent \textbf{Information Flow-Based Rewiring.}
In most graph mining algorithms, there are information flows from nodes to nodes~\cite{DBLP:conf/iclr/KipfW17,velivckovic2017graph,xu2018powerful,chen2009efficient}. An intuitive idea to mitigate group unfairness is to modify the graph topology to make such information flows as fair as possible.
%
%
For instance, Jalali et al.~\cite{jalali2020information} proposed \textit{Information Unfairness Score} based on the information flows. Specifically, given several groups of nodes, the information unfairness score depicts the largest distribution difference of the probabilistic accessibility between two node groups.
%
%
To obtain a fair graph topology, edges are rewired in a greedy manner to maximally reduce the information unfairness score.
Additionally, in GNNs, the information aggregation operation is found to introduce bias from the biased network topology to the learned node embeddings~\cite{dong2021edits,li2020dyadic,kose2022fair,kose2021fairness,jiang2022fmp}. Dong et al.~\cite{dong2021edits} proposed to perform edge rewiring for fair node embedding learning. 
Specifically, the Wasserstein distance between the node embedding distributions from two sensitive subgroups is minimized by learning a less biased (weighted) graph adjacency matrix. 
The learned weights in the adjacency matrix are converted into binary values according to a pre-assigned threshold for edge rewiring.
Similar edge rewiring ideas are also adopted by other works.
For example, Li et al.~\cite{li2020dyadic} proposed to optimize the adjacency matrix to minimize the expected probability difference of being connected between inter- and intra-group node pairs in link prediction tasks.

%

\noindent \textbf{Edge Sampling-Based Rewiring.}
Edges can also be sampled in a probabilistic way to improve group fairness. For example, Spinelli et al.~\cite{spinelli2021biased} pointed out that nodes within the same sensitive subgroup tend to be linked together on homogeneous graphs. The dominance of these intra-group edges could lead to bias in embedding learning. To tackle this issue, Spinelli et al. proposed a debiasing approach named FairDrop, where more intra-group edges than inter-group edges are removed according to a probabilistic edge-masking matrix. 
%
Similar probabilistic edge removing approaches are also adopted by other works such as~\cite{kose2022fair}.

\subsubsection{Improving Individual Fairness}

In terms of individual fairness, the edge rewiring strategy encourages similar individuals to share similar topological characteristics.
For example, algorithms based on a biased network topology tend to yield biased results in downstream tasks~\cite{laclau2021all,dong2021edits}. To tackle this issue, an edge rewiring strategy is introduced by Laclau et al.~\cite{laclau2021all} to achieve a fair topology for downstream tasks. Specifically, a matrix $\mathbf{S}$ is first given to indicate the pair-wise node similarity. To optimize the network topology, an optimization problem is then formulated to encourage similar nodes to have highly overlapped neighboring node sets after edge rewiring. Downstream tasks are proved to benefit from the rewired network topology in terms of individual fairness.


\subsubsection{Edge Rewiring in Applications}

\noindent \textbf{Recommender Systems.} In recommender systems, edge rewiring can be leveraged to tackle the well-known filter bubble problem.
For example, Masrour et al.~\cite{DBLP:conf/aaai/MasrourWYTE20} proposed an extended modularity score (as presented in Section \ref{modu}) of the graph as a popularity fairness indicator. 
Based on the obtained link prediction results, a proportion of links are rewired in a greedy manner to promote the modularity score,
which helps to achieve popularity fairness.

\noindent \textbf{Other Applications.} Edge rewiring is also proved to be effective in other real-world applications. For example, there are existing works achieving fair predictions on default and credit risks over bank clients networks~\cite{dong2021edits,loveland2022fairedit}.


\subsection{Orthogonal Projection}  
\begin{figure}[!t]
    \centering
    \includegraphics[width=0.48\textwidth]{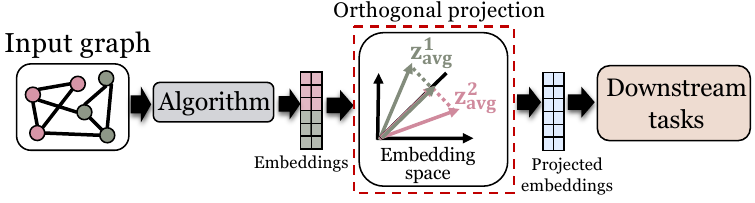}
    \vspace{-2mm}
    \caption{The pipeline of orthogonal projection-based approaches. The embeddings of all individuals are projected onto the same hyperplane in the embedding space.}
    \vspace{-5mm}
    \label{proj}
\end{figure}

Orthogonal projection is widely used for debiasing in the context of i.i.d. data to decorrelate the learned instance embeddings from sensitive attributes, e.g.,~\cite{sarhan2020fairness}.
Similarly, to decorrelate the embeddings of nodes from their sensitive feature(s), enforcing them to be orthogonal to the sensitive feature(s) is also an effective approach. This is usually achieved by projecting the node embeddings onto a hyperplane orthogonal to the direction of the sensitive features. We present a basic pipeline in Fig.~\ref{proj}. Compared with other debiasing techniques in graph mining, orthogonal projection provides a theoretical guarantee that node embeddings are uncorrelated with the sensitive feature(s)~\cite{DBLP:journals/corr/abs-1909-11793}.

\subsubsection{Improving Group Fairness}
Orthogonal projection is an effective approach to improve group fairness for graph embeddings.
In~\cite{zeng2021fair}, Zeng et al. defined \textit{Bias Direction}. Based on such, the node embeddings are projected onto a hyperplane orthogonal to the bias direction.
Specifically, for the $i$-th sensitive subgroup, we obtain an averaged unit node embedding as
\begin{align}
\mathbf{z}_{\text{avg}}^{i} = \frac{\mathbf{z}_1 + \mathbf{z}_2 + ... + \mathbf{z}_{|\mathcal{V}_i|}}{\| \mathbf{z}_1 + \mathbf{z}_2 + ... + \mathbf{z}_{|\mathcal{V}_i|} \|_2},
\end{align}
where for a binary sensitive feature, $i \in \{1,2\}$; $\mathbf{z}_j$ ($j \in \{1, 2, ...,|\mathcal{V}_i|\}$) denotes the learned embedding of node $v_j$. The unit vector in the bias direction is defined as 
\begin{align}
\mathbf{z}_{\text{bias}} = \frac{\mathbf{z}_{\text{avg}}^{1} - \mathbf{z}_{\text{avg}}^{2}}{\|\mathbf{z}_{\text{avg}}^{1} - \mathbf{z}_{\text{avg}}^{2}\|_2}.
\end{align}
Generally, if all node embeddings are projected onto a hyperplane that is orthogonal to $\mathbf{z}_{\text{bias}}$, then the component of the projected node embeddings in the direction of $\mathbf{z}_{\text{bias}}$ is zero, i.e., the sensitive information is decorrelated from the learned node embeddings.
Correspondingly, for node $v_j$, the projected embedding is formulated as $\mathbf{z}_j' = \mathbf{z}_j - <\mathbf{z}_j, \mathbf{z}_{\text{bias}}>\mathbf{z}_{\text{bias}}$,
%
where $\mathbf{z}_j'$ is the projected embedding for node $v_j$, and $<\cdot, \cdot>$ is the inner product operator.
%
%
Similarly, Palowitch et al.~\cite{DBLP:journals/corr/abs-1909-11793} proposed to learn topological embeddings by projecting the embedding onto a hyperplane orthogonal to the hyperplane of node features. This offers the theoretical guarantee that there will be no correlation between the potentially biased node features and topological embeddings. 
However, it is worth noting that orthogonal projection only guarantees that the embeddings and sensitive feature(s) are uncorrelated, while how to exclude non-linear dependency remains under-explored.


\subsection{Summary of Techniques for Improving Fairness} 




We provide a summary of the six discussed techniques for improving fairness. 
%
Optimization with regularization is the most widely used technique due to its simplicity and flexibility. Optimization with constraint(s) is less explored in deep learning based algorithms compared with those traditional ones, since adapting the constraint(s) for the gradient-based optimization can be difficult. The strategy of rebalancing is mostly designed to be tailored for specific application scenarios, where commonly used methods include up/down-sampling. Both adversarial learning and orthogonal projection aim to explicitly remove the sensitive information from the algorithm output. The former achieves such a goal via learning a discriminator, while the latter employs projection to remove the linear correlation between the model output and the sensitive attributes. We provide a detailed discussion on how well these techniques work based on the results reported in their papers in Appendix~\ref{how_well}.
\section{Research Challenges}
\label{challenge_and_lib}



Here we introduce the limitations of current research, pressing challenges, and open questions for future advances.

\noindent \textbf{Formulating Fairness Notions.}
Discrimination could exist in diverse forms in graph mining. Correspondingly,
different types of fairness notions should be formulated towards a comprehensive understanding of bias and discrimination in different real-world applications~\cite{caton2020fairness,burrell2016machine,skirpan2017authority,wong2020democratizing,yeung2018algorithmic}.
Although we have surveyed many fairness notions for graph mining, we need to admit that by no means are they complete, as other types of biases could also exist, depending on the needs of different real-world scenarios~\cite{xie2021fairrankvis}.
On the other hand, the definitions of different fairness notions on graphs could even be in conflict with each other~\cite{binns2020apparent,xie2021fairrankvis}. Therefore, designing a new fairness notion or choosing a set of existing non-conflicting fairness notions for particular graph mining algorithms and downstream applications remains an open question.



\noindent \textbf{Fulfilling Multiple Types of Fairness.}
It should be noted that any type of bias is undesired in real-world applications.
In this regard, there is an urgent need to promote multiple types of fairness at the same time. For example, group fairness and individual fairness can be promoted at the same time under certain scenarios~\cite{DBLP:journals/pvldb/LahotiGW19,fan2021fair}. However, promoting multiple types of fairness at the same time is a non-trivial problem, as promoting one type of fairness may degrade several other types of fairness~\cite{binns2020apparent,burkholder2021certification}.
Such a phenomenon can be more pronounced on graphs, which is resulted from the dependency between neighboring nodes.
%
For example, in a social network, individuals with the same gender are more densely connected. In this case, individual fairness enforces the nodes in the same gender subgroup to be similar (e.g., similar embeddings). However, such a goal may lead to a larger discrepancy between gender subgroups, which adversely affects the level of group fairness.
%
Therefore, properly addressing multiple unfairness issues in graph mining simultaneously is a pressing problem.

\noindent \textbf{Balancing Model Utility and Algorithmic Fairness.}
For algorithms with fairness considerations, the utility such as prediction accuracy is usually sacrificed~\cite{pannekoek2021investigating,xing2021fairness,DBLP:journals/corr/abs-2010-03240}. 
Such a trade-off between utility and fairness has been studied on i.i.d. data in recent years. To achieve a satisfying trade-off, a common strategy is to ensure the algorithm bearing Pareto optimality~\cite{martinez2019fairness,silvia2020general}, i.e., a state where either utility or fairness cannot be promoted without harming the other one.
Graph mining algorithms also have the issue of utility-fairness trade-off~\cite{ge2022toward,dai2021say,dong2021individual}. For example, when the fairness-related regularization is added to the objective function of a specific graph analytical task, the solution of the regularized optimization problem often deviates from the solution that brings the best utility in the unregularized optimization problem.
Additionally, in an adversarial learning-based framework, when the generator successfully fools the discriminator, some useful information may also be wiped out from the embeddings or predictions given by the generator. This could also degrade the model utility performance in downstream tasks. Hence it is critical to study how to achieve a trade-off between utility and fairness.

\noindent \textbf{Explaining How Unfairness Arises.}
Although various debiasing strategies have been proposed to debias graph mining algorithms, systematically understanding how such unfairness arises in the underlying algorithm is also crucial. However, this problem can be challenging. A reason is that the exhibited unfairness is usually coupled with both the input graph and specific mechanisms in graph mining algorithms. For example, due to the message-passing mechanism in GNNs, the unfairness exhibited in the learned node embeddings can be attributed to the biased input graph topology~\cite{dong2021edits}. Systematically explaining how unfairness arises in various graph mining algorithms remains a critical issue to be addressed.

\noindent \textbf{Enhancing Robustness of Algorithms on Fairness.}
In graph mining, enhancing the robustness of graph mining algorithms w.r.t. fairness is another urgent need. For instance, in learning-based algorithms, human annotators could provide biased supervision information for model training~\cite{buolamwini2018gender}. Besides, the algorithms may also be manipulated by malicious attackers to exhibit discrimination against a certain group of people~\cite{solans2020poisoning,angwin2016machine}.
In both cases, the fairness level of the algorithm predictions can be dramatically lowered.
Despite the significance of enhancing the robustness of algorithmic fairness, most existing studies are overwhelmingly devoted to i.i.d. data~\cite{solans2020poisoning,mehrabi2020exacerbating}, and cannot be directly grafted to the graph-structured data.
In this regard, how to promote the robustness of the fairness aspect of graph mining algorithms deserves further investigation.

\section{Conclusion}

\label{conclusion_section}

Graph mining has achieved remarkable success in a myriad of high-impact real-world applications. Nevertheless, due to the lack of fairness considerations, there has been an increasing societal concern that these algorithms may exhibit discrimination when they are exploited to make predictions and decisions. Over the years, many efforts have been made to define, measure, and promote fairness in graph mining. In this survey, we propose a novel taxonomy of fairness notions in graph mining research and systematically review existing fairness notions from different perspectives. Besides, we categorize and introduce existing techniques that promote fairness in graph mining. Furthermore, rich benchmark graph datasets are collected to facilitate future research advances in this area. Finally, existing challenges and open questions areas are also discussed.

\section{Acknowledgements}

This work is supported by the National Science Foundation under grants IIS-2006844, IIS-2144209, IIS-2223769, CNS-2154962, and BCS-2228534, the JP Morgan Chase Faculty Research Award, the Cisco Faculty Research Award, and Jefferson Lab subcontract JSA-22-D0311.

\bibliographystyle{plain}
\bibliography{ref}

\vspace{-10mm}
\begin{IEEEbiography}[{\includegraphics[width=1in,height=1.25in,clip,keepaspectratio]{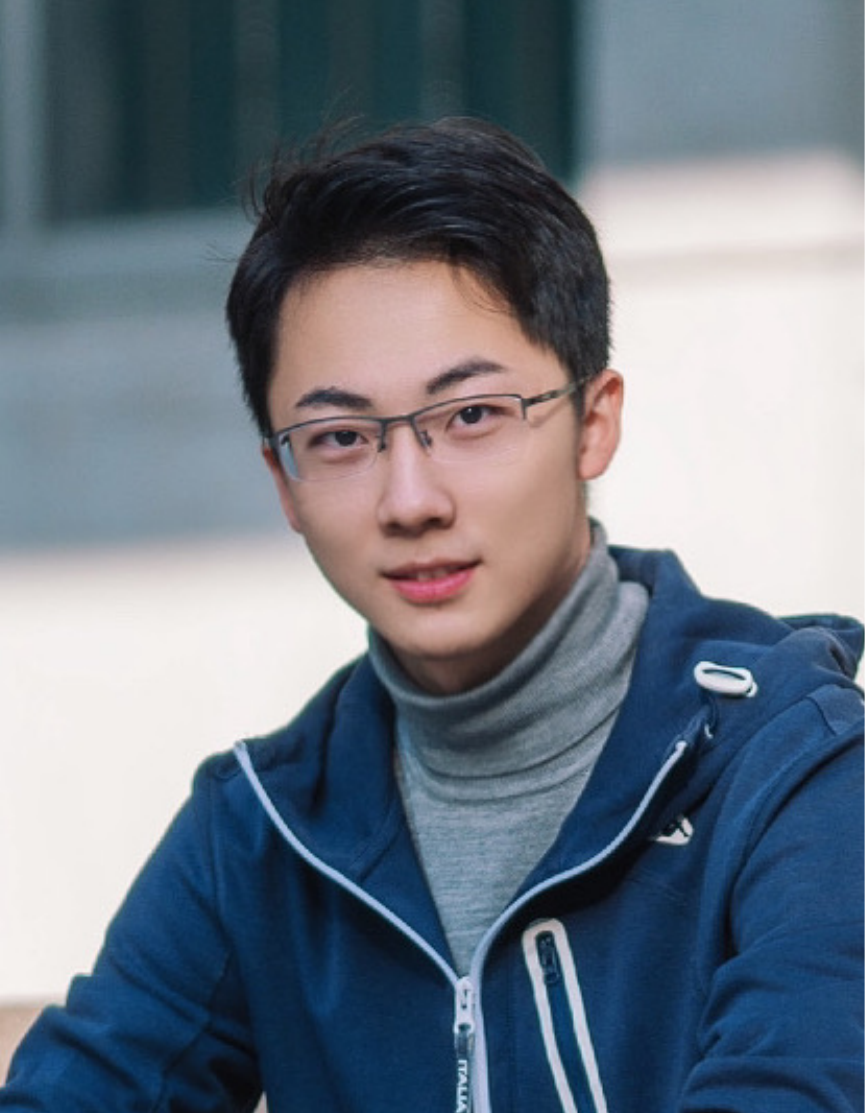}}]{Yushun Dong}
Yushun Dong is a Ph.D. student in the Department of Electrical and Computer Engineering at the University of Virginia. He received the B.S. degree in Telecommunications from Beijing University of Posts and Telecommunications in 2019. His research interests are broadly in data mining and machine learning, with a particular focus on graph learning algorithms. In the past few years, his works have been published in top-tier venues including SIGKDD, WWW, and CIKM.
\end{IEEEbiography}
\vspace{-8mm}
\begin{IEEEbiography}[{\includegraphics[width=1in,height=1.25in,clip,keepaspectratio]{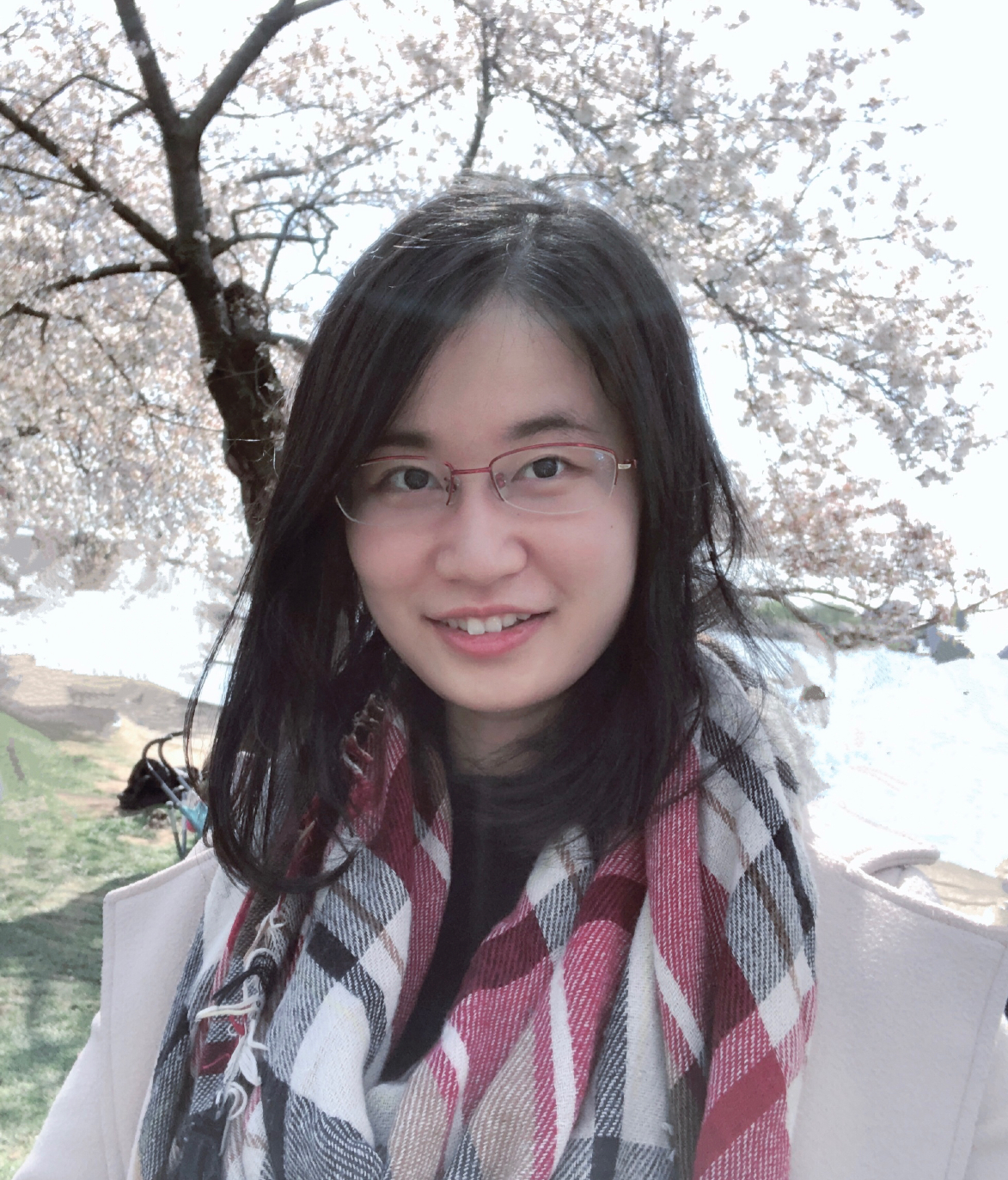}}]{Jing Ma}
Jing Ma is a Ph.D. candidate in the Department of Computer Science at the University of Virginia. Her research interests include causal inference, machine learning, data mining, and especially for bridging the gap between causality and machine learning. Her works have been published in top conferences and journals such as IJCAI, WWW, AAAI, TKDE, WSDM, SIGIR, and IPSN.
\end{IEEEbiography}
\vspace{-8mm}
\begin{IEEEbiography}[{\includegraphics[width=1in,height=1.25in,clip,keepaspectratio]{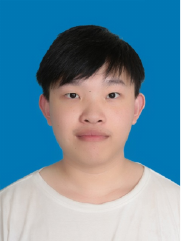}}]{Song Wang}
is a Ph.D. student at the Department of Electrical and Computer Engineering at the University of Virginia, advised by Professor Jundong Li. Previously he received a B.E. degree in Electronic Engineering from Tsinghua University in 2020. His research interests include knowledge graphs and few-shot learning on graphs. His works have been published in top conferences such as IJCAI, NeurIPS, WSDM, SIGIR, and SIGKDD.
\end{IEEEbiography}
\vspace{-8mm}
\begin{IEEEbiography}[{\includegraphics[width=1in,height=1.25in,clip,keepaspectratio]{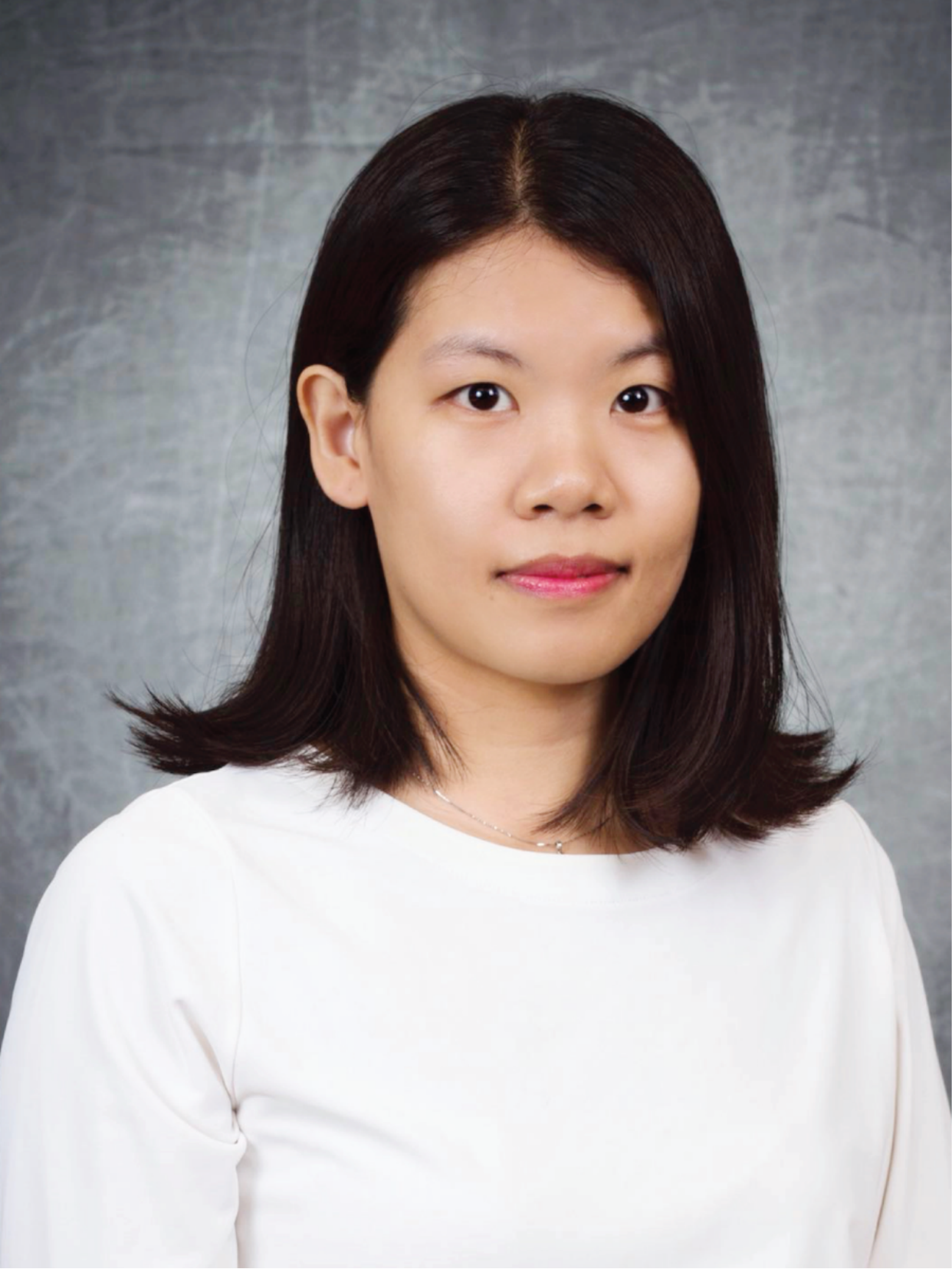}}]{Chen Chen}
Chen Chen is a Research Assistant Professor at the University of Virginia. Before joining the University of Virginia, Chen was a software engineer at Google working on personalized recommendations for Google Assistant. Chen got her Ph.D. degree from Arizona State University. Her research focuses on the connectivity of complex networks and has been applied to address pressing challenges in various high-impact domains, including social media, bioinformatics, recommendation, and critical infrastructure systems. Her research has appeared in top-tier conferences (including KDD, ICDM, SDM, WSDM, SIGIR, AAAI, and IJCAI), and prestigious journals (including IEEE TKDE, ACM TKDD, and SIAM SAM). Chen has received several awards, including Bests of SDM'15, Bests of KDD'16, Rising Star in EECS'19, Outstanding Reviewer of WSDM'21.
\end{IEEEbiography}
\vspace{-8mm}
\begin{IEEEbiography}[{\includegraphics[width=1in,height=1.25in,clip,keepaspectratio]{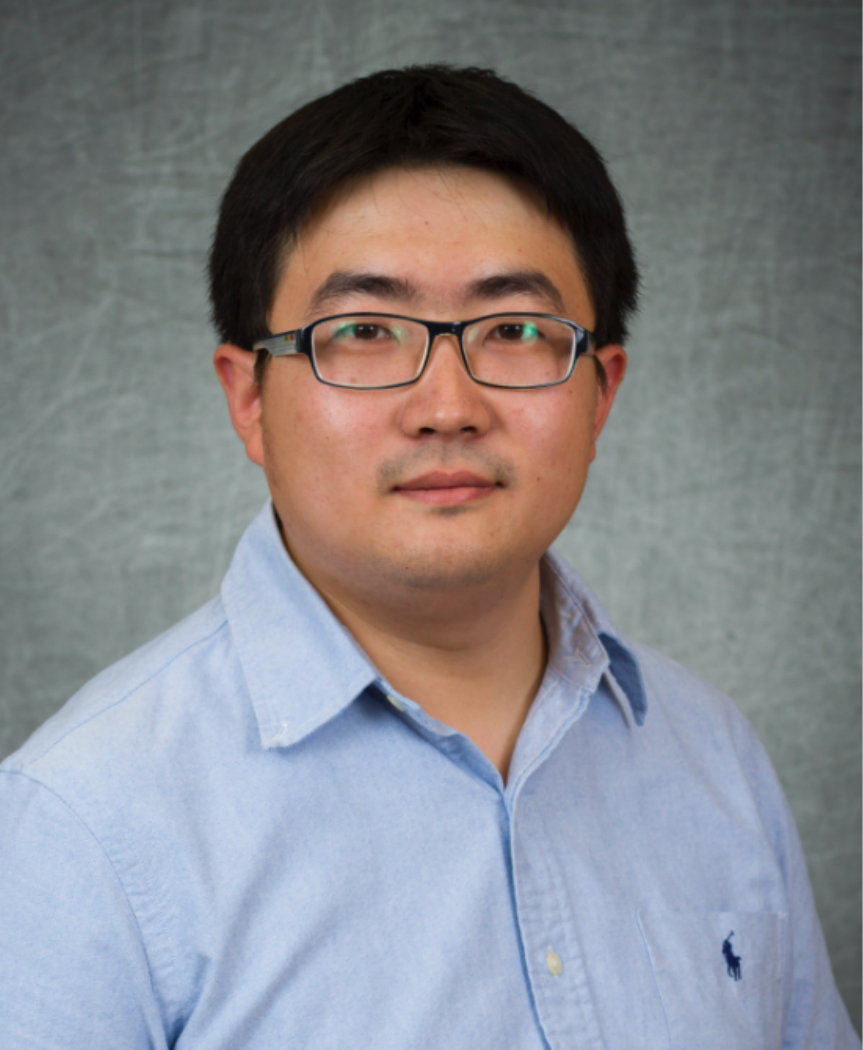}}]{Jundong Li}
Jundong Li is an Assistant Professor in the Department of Electrical and Computer Engineering, with a joint appointment in the Department of Computer Science, and the School of Data Science. He received Ph.D. degree in Computer Science at Arizona State University in 2019. His research interests are in data mining, machine learning, and causal inference. He has published over 120 articles in high-impact venues (e.g., KDD, WWW, NeurIPS, AAAI, IJCAI, WSDM, EMNLP, CSUR, TPAMI, TKDE, TKDD, and TIST). He has won several prestigious awards, including NSF CAREER Award, KDD Best Research Paper Award, JP Morgan Chase Faculty Research Award, Cisco Faculty Research Award, and being selected for the AAAI 2021 New Faculty Highlights program.
\end{IEEEbiography}

\clearpage

\appendices

\linespread{1.01}

\section{Techniques Comparison Under Same Fairness Notions}
\label{comparison_under_same_notion}

\subsection{Group Fairness}

\noindent Multiple techniques are utilized to fulfill group fairness according to the surveyed literature, including optimization with regularization, optimization with constraint(s), rebalancing, adversarial learning, edge rewiring, and orthogonal projection. These techniques fulfill group fairness in different ways. More specifically, optimization with regularization and adversarial learning directly enforce a fairness-aware goal for the optimization of graph mining algorithm: regularization terms serve as an explicit fairness-aware objective added onto the overall objective function, while adversarial learning aims to obtain prediction results that fool a discriminator (on the information about sensitive attributes). Different from the two techniques above, optimization with constraint(s) directly defines a fair feasible set under a certain group fairness notion, and optimization results can only be within such a region. Rebalancing achieves debiasing in another way: this technique aims to modify the graph mining algorithm in most cases, which empirically leads to less statistical prediction differences (e.g., positive rate) among demographic subgroups. Usually, such statistical prediction differences are closely related to certain fairness notions, and thus the graph mining algorithm is debiased.
Finally, in the surveyed literature, both edge rewiring and orthogonal projection are applied to modify the predictions of graph mining algorithms to fulfill group fairness: edge rewiring is usually adopted to modify the link prediction results, while orthogonal projection directly achieves debiasing in the embedding space and thus is suitable for most node embedding learning tasks.

\subsection{Individual Fairness}

\noindent Optimization with regularization, optimization with constraint(s), and edge rewiring are surveyed techniques that can be applied to fulfill individual fairness. Among them, optimization with regularization aims to penalize the exhibited bias under a certain notion of individual fairness. Optimization with constraint(s) requires first defining a feasible set that can be considered fair according to a certain notion of individual fairness. Different from them, edge rewiring directly modifies the predicted links to mitigate the exhibited individual-level bias.

\subsection{Fairness in Recommender Systems}

\noindent Optimization with regularization, rebalancing, and edge rewiring are adopted in the surveyed literature to fulfill fairness in recommender systems. The primary goal of optimization with regularization is to penalize the exhibited unfairness during optimization (similar to its goal under group fairness and individual fairness). Rebalancing here focuses on the input data in most cases, aiming to ensure a balanced appearance rate for instances (e.g., users and items) from different groups (e.g., gender groups of users and provider groups of items). Finally, edge rewiring serves as a post-processing approach, which modifies the predicted connections between recommendations and users to mitigate bias (similar to its goal under individual fairness).

\subsection{Fairness in Knowledge Graphs}

\noindent Both optimization with regularization and adversarial learning can be applied to fulfill fairness in knowledge graphs. Here both approaches share a similar goal of enforcing a fairness-aware objective function during the optimization of the graph mining algorithm. The main difference is that optimization with regularization requires formulating such a goal mathematically according to certain fairness notions, while in adversarial learning, the fairness-aware goal is formulated as the performance of a discriminator on identifying the information about bias.

\section{Effectiveness of the surveyed techniques}
\label{how_well}

We then discuss the effectiveness of the surveyed techniques, i.e., how well these techniques work based on the results reported in their vanilla papers. We provide a discussion for each of the six mainstreams below.

\noindent \textbf{Optimization with Regularization.} Designing regularization term(s) to improve fairness is the most widely used strategy due to its simplicity and flexibility. Nevertheless, its effectiveness may largely vary under different datasets and backbone objective functions or models~\cite{DBLP:conf/sigir/FuXGZHGXGSZM20,fan2021fair}. For example, the effectiveness of optimization with regularization in improving fairness is limited~\cite{DBLP:journals/pvldb/LahotiGW19,fan2021fair,dong2021individual} or even counterproductive~\cite{agarwal2021towards} in certain cases based on the results reported in multiple surveyed papers. Moreover, the hyper-parameter tuning can also influence its effectiveness~\cite{DBLP:conf/sigir/ZhuWC20}.

\noindent \textbf{Optimization with Constraint(s).} Optimization with constraint(s) is mostly adopted in traditional graph mining algorithms instead of those based on deep learning, since it is difficult to incorporate the constraint(s) into the gradient-based optimization process. Based on the reported results in the surveyed papers, we found that despite its effectiveness in improving fairness, the interest of the advantaged group could be largely reduced~\cite{ali2019fairness}. In addition, the overall algorithmic utility may also be jeopardized by a large margin~\cite{farnad2020unifying,DBLP:conf/icml/KleindessnerSAM19}. The reason is that it is difficult to design proper constraint(s) that ensure that the feasible set contains solution points with a high level of utility.

\noindent \textbf{Rebalancing.} The rebalancing strategy is mostly designed to be tailored for specific application scenarios. Although its effectiveness in improving fairness has been widely acknowledged, we found that it also bears disadvantages based on the reported results in the surveyed papers. For example, rebalancing may result in large fairness level variances in certain cases~\cite{DBLP:conf/icml/BuylB20}. In addition, techniques based on rebalancing could be less competitive~\cite{DBLP:journals/corr/abs-1809-09030,DBLP:journals/corr/abs-2105-02725} or more limited by the application scenarios compared with other existing baselines~\cite{DBLP:journals/corr/abs-2006-04279}.

\noindent \textbf{Adversarial Learning.} Based on the reported results in the surveyed papers, adversarial learning is also an effective strategy to achieve higher levels of algorithmic fairness~\cite{dai2021say,krishnan2018adversarial}. However, we found that it can be harder to balance the prediction utility and fairness for techniques based on adversarial learning~\cite{DBLP:conf/icml/BoseH19}, the prediction utility could be clearly jeopardized in certain cases~\cite{xu2021fair}.

\noindent \textbf{Edge Rewiring.} Edge rewiring is less explored compared with other popular approaches. A preliminary reason is that it is difficult to adapt those highly efficient gradient-based optimization techniques to obtain the optimal edge rewiring solution(s) due to the binary nature of most graphs in real world. Based on the reported results in most surveyed papers, edge rewiring is able to provide stable improvement in the prediction fairness and maintain the utility at the same time~\cite{li2020dyadic}. Nevertheless, it could be not competitive enough compared with baselines in certain cases~\cite{dong2021edits,spinelli2021biased,laclau2021all}, and further improvements in their effectiveness of achieving a higher level of fairness may still worth exploring.

\noindent \textbf{Orthogonal Projection.} It is also proved that orthogonal projection can also achieve a high level of fairness in graph mining~\cite{palowitchdebiasing,xu2021fair}. Nevertheless, most existing explorations are based on the linear assumption between the sensitive attributes and predictions~\cite{palowitchdebiasing}. Whether non-linear dependencies jeopardize the level of fairness with such an approach is still under-explored.

\begin{table*}[!t]  
\caption{Collection of benchmark datasets. ``CI, C2, C3, C4" refer to social networks, recommendation-based networks, academic networks, and other types of networks, respectively. Different versions exist for datasets marked with ``$^*$", and only the statistics corresponding to the most representative version are presented.}
\small
\label{tab:dataset}
\small
\setlength\tabcolsep{1.4pt}  
\renewcommand{\arraystretch}{1.0}  
\begin{tabular}{l|llccccc}
\hline
\hline
                      & \textbf{Dataset}                             & \textbf{Fairness Goal(s)}                                                                                                & \textbf{\# Nodes}                & \textbf{\# Edges}                   & \textbf{\# Features} & {$\bm{S}$ $\bm{(|S|)}$}   & \textbf{Works } \\
\hline
\hline
\multirow{11}{*}{\textbf{C1}}          & \multirow{2}{*}{Facebook$^*$ \cite{he2016ups}}                           & \multirow{2}{*}{\makecell[l]{group, \\individual}}                                                                                                 &   \multirow{2}{*}{1,034}  & \multirow{2}{*}{{26,749}}    &    \multirow{2}{*}{224}       &   \multirow{2}{*}{gender (2)}                                &     \cite{ali2019fairness,DBLP:conf/pkdd/BuylB21,fan2021fair,current2022fairmod,DBLP:conf/icml/KleindessnerSAM19,dong2021individual,DBLP:conf/kdd/KangHMT20,DBLP:conf/ijcai/KhajehnejadRBHJ20}            \\
                              &&&&&&&   \cite{saxena2021hm,krasanakis2020applying,zeng2021fair,teng2021influencing,laclau2021all,DBLP:conf/aaai/MasrourWYTE20}\\
\cline{2-8}
 & Pokec$^*$ \cite{takac2012data}     & group                                 & 1,632,803               & 30,622,564              & {59} & region (2), gender (2)  & \cite{dai2021say,navarin2020learning,franco2022deep,kose2022fair,jiang2022fmp,dong2021edits} \\
\cline{2-8}
  & Twitter$^*$    \cite{mcauley2012learning}                         & group                                                                                                 &  {81,306} & {1,768,149} &   1,364          &                political opinion  (2)                   &       \cite{krasanakis2020applying,tsioutsiouliklis2021fairness,DBLP:journals/corr/abs-2105-02725}         \\ 
  \cline{2-8}
 & Lastfm \cite{cantador2011second}     & \makecell[l]{group, \\provider}                                                                 &     49,900            &      518,647                      &        -                       &          gender (2), age (3)                                          &      \cite{patro2020fairrec,xu2021fair,wu2021learning}          \\
\cline{2-8}
& Ok97 \cite{red2011comparing} & group      &3,111                   &      73,230                                                &  8              &            gender (2)
                                   &  \cite{li2020dyadic}              \\
                                   \cline{2-8}
& UNC28 \cite{red2011comparing} & group & 4,018 & 65,287 &   8        &         gender (2)
                               &  \cite{li2020dyadic}              \\
                               \cline{2-8}
 & Google+   \cite{mcauley2012learning}                          & popularity                              &   4,938                                          &     547,923                          &     5           &       -                          &   \cite{DBLP:conf/aaai/MasrourWYTE20}                \\
                              \cline{2-8}
 & Epinion \cite{tang2012mtrust}  & popularity                            & 8,806 & 157,887 &        -             & - &                 \cite{zhu2021popularity,abdollahpouri2017controlling} \\
 \cline{2-8}
 & Filmtrust \cite{guo2013novel} & provider & 3,579                  & 35,494                &        -              &    -         &   \cite{liu2018personalizing}\\
 \cline{2-8}
 & Ciao \cite{tang2012mtrust}     & popularity                            & 7,317                & 85,205               & -   &       -              & \cite{zhu2021popularity}\\
\hline
\multirow{8}{*}{\textbf{C2}} & Amazon$^*$  \cite{leskovec2007dynamics}                            & \makecell[l]{group, \\marketing, \\social}                                                                                    &  334,863               &   925,872                          & -                              &       product category  (4)                                                        &                         \cite{DBLP:conf/sigir/ZhuWC20,DBLP:conf/sigir/FuXGZHGXGSZM20,krasanakis2020applying,li2021user,DBLP:conf/wsdm/WanNMM20}              \\
\cline{2-8}
                              & Yelp$^*$~\cite{DBLP:conf/sigir/ZhuWC20}                              & \makecell[l]{group, \\social}                                                                                    &      12,683           &    211,721                         &    14                &                                                 food genre   (4)                                        &       \cite{DBLP:conf/sigir/ZhuWC20}         \\
                              \cline{2-8}
                              & \multirow{2}{*}{ML100K \cite{harper2015movielens} }                     & \multirow{2}{*}{\makecell[l]{group,\\ popularity}}                            &   \multirow{2}{*}{2,625}      &   \multirow{2}{*}{100,000}                            &      \multirow{2}{*}{12}          &                       gender (2), age (7),                                      &  \multirow{2}{*}{\cite{DBLP:journals/corr/abs-2101-03584,DBLP:conf/icml/BuylB20,DBLP:conf/pkdd/BuylB21,DBLP:journals/corr/abs-1909-11793}}     \\
                              &&&&&&occupation (21) &\\
                              \cline{2-8}
                              & \multirow{3}{*}{ML1M \cite{harper2015movielens}}& \makecell[l]{group, \\individual,}  &     \multirow{3}{*}{10,000}            &        \multirow{3}{*}{1,000,000}                    &      \multirow{3}{*}{11}                         &     \multirow{2}{*}{ gender (2), age (7), }              &   \cite{abdollahpouri2017controlling,DBLP:journals/corr/abs-2101-03584,burke2017balanced,DBLP:conf/icml/BoseH19,DBLP:conf/sigir/ChenXLYSD20,DBLP:journals/corr/abs-1809-09030}         \\         
      && \makecell[l]{popularity, \\ provider,}&&&&\multirow{2}{*}{occupation (21)} &\cite{kamishima2017considerations,DBLP:conf/recsys/KayaBT20,DBLP:conf/recsys/LinZZGLM17,DBLP:conf/um/MalecekP21,liu2018personalizing}                          \\
      &&\makecell[l]{social, \\user}&&&&& \cite{wu2021learning,zeng2021fair,zhu2021popularity,xu2021fair,DBLP:conf/sigir/ZhuWC20}\\
      \cline{2-8}
                              & ML20M   \cite{harper2015movielens}                    & popularity                                                                                            &      165,000           &  20,000,000                          &     6                          &       -         &                                             \cite{krishnan2018adversarial,DBLP:conf/flairs/WasilewskiH16}             \\
\hline
\multirow{5}{*}{\textbf{C3}}             & Citeseer \cite{sen2008collective}                            & \makecell[l]{group, \\degree-related}    &        3,327         &       4,732                     &    3,703                           &                                              topic (6)            &     \cite{tang2020investigating,kose2022fair,current2022fairmod,li2020dyadic,spinelli2021biased}           \\
\cline{2-8}
                              & Cora   \cite{sen2008collective}                              & \makecell[l]{group,\\ degree-related}                                                                                            &      2,708           &  5,429                          &   1,433                            &                                            topic (7)                                &     \cite{tang2020investigating,kose2022fair,current2022fairmod,li2020dyadic,spinelli2021biased}           \\
                              \cline{2-8}
                              & Pubmed   \cite{sen2008collective}                            & \makecell[l]{group,\\ degree-related}                                                                                            &  19,717               &   44,338                         &       500                        &                                           topic (3)                       &    \cite{tang2020investigating,kose2022fair,current2022fairmod,li2020dyadic,spinelli2021biased}            \\
                              \cline{2-8}
                              & \multirow{2}{*}{DBLP$^*$  \cite{tang2008arnetminer}}                              & \multirow{2}{*}{\makecell[l]{group, \\individual}}                                     &   \multirow{2}{*}{3,980}               &  \multirow{2}{*}{6,965}                           &  \multirow{2}{*}{-}    &   continent (5),                                                                               & \multirow{2}{*}{\cite{DBLP:conf/icml/BuylB20,laclau2021all,jalali2020information,stoica2020seeding,tsioutsiouliklis2021fairness,spinelli2021biased}}               \\
                              && &&&& gender (2) &\\
\hline
\multirow{8}{*}{\textbf{C4}} & \multirow{2}{*}{German \cite{dua2017uci}}    & \multirow{2}{*}{\makecell[l]{group, \\ couterfactual}}                          & \multirow{2}{*}{1,000}  & \multirow{2}{*}{21,742}    & \multirow{2}{*}{27} & \multirow{2}{*}{gender (2)}   & \multirow{2}{*}{\cite{zhang2021multi,agarwal2021towards,dong2021edits}} \\
 &&&&&&&\\
 \cline{2-8}
       & NBA~\cite{dai2021say}      & group                               & 403    & 10,621    & 96 & country (2) & \cite{dai2021say,jiang2022fmp}  \\
       \cline{2-8}
       & \multirow{2}{*}{Recidivism \cite{jordan2015effect}}& \makecell[l]{group, \\ individual,} & \multirow{2}{*}{18,876} & \multirow{2}{*}{311,870}   & \multirow{2}{*}{18} & \multirow{2}{*}{race (2)}  &  \multirow{2}{*}{\cite{zhang2021multi,agarwal2021towards,ma2022Learning,fan2021fair,dong2021edits}}\\
       &&couterfactual&&&&&\\
       \cline{2-8}
       & \multirow{2}{*}{Credit \cite{yeh2009comparisons}}    & \multirow{2}{*}{\makecell[l]{group, \\couterfactual}}                     &\multirow{2}{*}{30,000} & \multirow{2}{*}{1,421,858} & \multirow{2}{*}{13} & \multirow{2}{*}{age (2)}  & \multirow{2}{*}{\cite{zhang2021multi,agarwal2021towards,ma2022Learning,dong2021edits}}\\
       && &&&&&\\
\hline
\hline
\end{tabular}
\end{table*}

\section{Online Resources}
\label{dataset_section}

\noindent \textbf{Benchmark Datasets.}
We summarize the datasets in Table~\ref{tab:dataset}. In general, we group these datasets into four categories w.r.t. the type of networks, including social networks (C1), recommendation-based networks (C2), academic networks (C3), and other types of networks (C4).

\noindent \textbf{Popular Algorithms.}
To help researchers and engineers compare and select appropriate graph mining algorithms with fairness considerations, we developed an open-source library \textit{PyGDebias}, which could facilitate the usage of different debiasing techniques in practice.

The benchmark datasets, fairness-aware graph mining algorithm implementations, and performance leaderboards over different fairness notions can be found in \url{https://github.com/yushundong/PyGDebias}.

\end{document}